\documentclass[10pt,journal]{IEEEtran}

\usepackage[normalem]{ulem}

\usepackage[utf8]{inputenc}
\usepackage[english]{babel}
\usepackage{csquotes}

\usepackage{multirow}
\usepackage{diagbox}
\usepackage{adjustbox}
\usepackage{soul, color}

\graphicspath{{figs/}{../jpeg/}}
\DeclareGraphicsExtensions{.pdf,.jpeg,.png}

\usepackage{amsmath}
\usepackage{amsfonts}
\usepackage{amssymb}

\usepackage[caption=false,font=footnotesize]{subfig}
\usepackage{graphicx}

\usepackage{url}

\usepackage[style=numeric,backend=biber]{biblatex}

\addto\captionsenglish{}
\addto\captionsenglish{}

\bibliography{biblio} 

\begin{document}
%
\title{Examining Deep Learning Architectures for Crime Classification and Prediction}

%
%
%

\author{Panagiotis~Stalidis,
  Theodoros~Semertzidis,~\IEEEmembership{Member,~IEEE} and
  Petros~Daras,~\IEEEmembership{Senior~Member,~IEEE}
\thanks{P. Stalidis, T. Semertzidis and P. Daras are with the Information Technologies Institute, Centre for Research and Technology Hellas, Thessaloniki, Greece. Email: {stalidis,theosem,daras}@iti.gr}
\thanks{Manuscript received 19 Sep. 2017; revised 22 Mar. 2018}
}

\IEEEtitleabstractindextext{%
\begin{abstract}
In this paper, a detailed study on crime classification and prediction using deep learning architectures is presented.
We examine the effectiveness of deep learning algorithms on this domain and provide recommendations for designing and training deep learning systems for predicting crime areas, using open data from police reports.
Having as training data time-series of crime types per location, a comparative study of 10 state-of-the-art methods against 3 different deep learning configurations is conducted. In our experiments with five publicly available datasets, we demonstrate that the deep learning-based methods consistently outperform the existing best-performing methods. Moreover, we evaluate the effectiveness of different parameters in the deep learning architectures and give insights for configuring them in order to achieve improved performance in crime classification and finally crime prediction.
\end{abstract}

}

\maketitle

\IEEEdisplaynontitleabstractindextext

%
\IEEEpeerreviewmaketitle

\section{Introduction}\label{sec:introduction}

%
%
%
%

\IEEEPARstart{P}{redictive} policing is the use of analytical techniques to identify either likely places of future crime scenes or past crime perpetrators, by applying statistical predictions \cite{perry2013predictive}. As a crime typically involves a perpetrator and a target and occurs at a certain place and time, techniques of predictive policing need to answer: a) who will commit a crime, b) who will be offended, c) what type of crime, d) in which location and e) at what time a new crime will take place. This work does not focus on the victim and the offender, but on the prediction of occurrence of a certain crime type per location and time using past data.
 
 The ultimate goal, in a policing context, is the selection of the top areas in the city for the prioritization of law enforcement resources per department. 
One of the most challenging issues of police departments is to have accurate crime forecasts to dynamically deploy patrols and other resources so as to improve deterring of crime occurrence and police response times. 

Routine activity theory \cite{cohen1979social} suggests that most crimes take place when three conditions are met: a motivated offender, a suitable victim and lack of victim protection. The rational choice theory \cite{cornish2014reasoning}, suggests that prospective criminal weights the gain of successfully committing the crime against the probability of being caught and makes a rational choice whether to actually commit the crime or not. Both theories agree that a crime takes place when a person willing to commit it has an opportunity to do so. As empirical studies in near repeat victimization 
\cite{grubesic2008spatio, johnson1997new, johnson2007space, bowers2004commits} 
have shown, these opportunities are not randomly distributed, but follow patterns in both space and time. Traditionally, police officers use maps of an area and place a pin on the map for every reported incident. Studying these maps, they can detect these patterns and thus, to efficiently predict hotspots; A hotspot is defined as the area with the higher possibility for a crime to occur, compared to the neighbouring areas.

Simple mapping methods are not sufficient to make use of these general phenomena as early indicators for predicting crimes but more complex methodologies, such as machine learning, are needed. Various machine learning methodologies like Random Forests \cite{breiman2001random}, Naive Bayes \cite{zhang2004optimality} and Support Vector Machines (SVMs) \cite{cortes1995support} have been exploited in the literature both for predicting the number of crimes that will occur in an area and for hotspot prediction. The success of a machine learning analysis highly depends on the experience of the analyst to prepare the data and to hand-craft features that describe properly the problem in question. 

Deep learning is a machine learning approach where the algorithm can extract the features from the raw data, overcoming the limitations of other machine learning methodologies. Of cource this benefit comes at a high price in computational complexity and demand in raw data. This global research trend was picked recently by Wang \textit{et al.} \cite{wang2017deep} for predicting hourly fluctuations in crime rates. Building on this very interesting work, we investigate DL architectures for crime hotspot prediction and present design recommendations. The major contributions of this paper are:

\begin{itemize}
\item We present 3 fundamental DL architecture configurations for crime prediction based on encoding: a) the spatial and then the temporal patterns, b)  the temporal and then the spatial patterns, c) temporal and spatial patterns in parallel. 
\item We experimentally evaluate and select the most efficient configuration to deepen our investigation. 
\item We compare our models with 10 state-of-the-art algorithms on 5 different crime prediction datasets with more than 10 years of crime report data.
\item Finally, we propose a guide for designing DL models for crime hotspot prediction and classification. 
\end{itemize}

The rest of the paper is organized as follows: Section \ref{relatedWork}, discusses the related work on crime prediction and classification as well as recent developments in deep learning approaches for spatio-temporal data. Section \ref{problemFormulation}, formulates the problem in question. Section \ref{methodology}, discusses the proposed methods for applying DL in crime prediction and classification. Section \ref{experimentalSetup} presents the baseline approaches, the datasets used and the metrics to measure the effectiveness of each model. In Section \ref{results}, the presentation and discussion of results in different configurations and 

\section{Related Work} \label{relatedWork}

Criminology literature investigates the relationship between crime and various features, developing approaches for crime forecasting. The majority of the works focus on the prediction of hotspots, which are areas of varying geographical size with high crime probability. The methods include Spatial and Temporal Analysis of Crime (STAC)\cite{levine2002spatial}, Thematic Mapping \cite{williamson2001tools} and Kernel Density Estimation (KDE) \cite{rosenblatt1956remarks}.

In STAC, the densest concentrations of points on the map are detected and then fit to a standard deviational ellipse for each one. Through the study of the size and the alignment of the ellipses, the analyst can draw conclusions about the nature of the underlying crime clusters \cite{chainey2008utility}.

In Thematic Mapping, the map is split in boundary areas while offences are placed as points on a map. The points can then be aggregated to geographic unit areas and shaded in accordance with the number of crimes that fall within \cite{williamson2001tools}. This technique enables quick determination of areas with a high incidence of crime and allows further analysis of the problem by “zooming in” on those areas. Boundary areas can be arbitrarily defined, using i.e. police beats, enabling linking of crime with other data sources, such as population.

Owing to the varying size and shape of most geographical boundaries, thematic mapping can be misleading in identifying the existence of the highest crime concentrations \cite{eck2005mapping}. Hence, this technique can fail to reveal patterns across and within the geographical division of boundary areas \cite{chainey2013gis}. KDE divides the area in a regular grid of cells and estimates a density value for each cell, using a kernel function that estimates the probability density of the actual crime incidents \cite{bishop2006pattern, silverman1986density}. The resulting two-dimensional scalar can then be used to create a heatmap, affected by the cell size, the bandwidth and the kernel function used \cite{de2016mskde}.

All three abovementioned methods solely rely on the spatial dimension of 
incidents. On the other hand, methods presented by Mohler \textit{et al.} \cite{mohler2011self} and Ratcliffe \cite{ratcliffe2006temporal}, model the temporal dimension of the crime. Mohler \textit{et al.} propose the use of self-exciting point process to model the crime and gain insights into the temporal trends in the rate of burglary, while  Ratcliffe investigates the temporal constraints on crime and proposes an offender travel and opportunity model. These works validate the claim that a proportion of offending is driven by the availability of opportunities presented in Cohen's routine activity theory \cite{cohen1979social}.

Nakaya and Yano \cite{nakaya2010visualising}, extend the crime cluster analysis with a temporal dimension. They employ the space-time variants of KDE to simultaneously visualize geographical extent and duration of crime clusters.
Taking a step further, Toole \textit{et al.} \cite{toole2011spatiotemporal}, use criminal offense records to identify spatio-temporal patterns at multiple scales. They employ various quantitative tools from mathematics and physics and identify significant correlation in both space and time in the crime behavioral data. 

Machine learning has also been a popular approach for crime forecasting. Olligschlaeger \cite{olligschlaeger1997artificial} examines the use of Multi-Layer Perceptrons (MLP) on GIS systems. One of the use cases is the prediction of drug related calls for service on the 911 call centers of Pittsburgh USA. By super imposing the map area with a grid of cells, Olligschlaeger creates 445 cells of a 2150 sq feet area each. For each cell, 3 call related early indicators are calculated: a) the number of weapon related calls, b) the number of robbery related calls and c) the number of  assaults related calls that occur in the cell area. Additionally, the proportion of commercial to residential properties in the cell area and a seasonal index are also used as indicators. Due to the lack of processing power in 1997, the MLP neural network that was used had a mere 9 neurons in a single hidden layer.

Kianhmer and Alhajj \cite{kianmehr2008effectiveness}, use SVMs in a machine learning approach of hotspot location prediction. The success of SVMs is re-examined by Yu \textit{et al.} \cite{yu2011crime} in comparison to other machine learning approaches like Naive Bayes and Random Forests. They observe that
in the case of residential burglary, what has happened in a particular place is likely to reoccur.

Gorr and Olligschlaeger compare different regression approaches for predicting a set of crime categories using data from Pittsburgh \cite{gorr2003short}. They run regressions of different complexity on the same data set and compare the results. They found that simple time series were outperformed by more sophisticated methods. In particular, they found that by using a smoothing coefficient (i.e. applying increased weight on recent data) the predicted mean absolute percent error is improved.


Xu \textit{et al.} \cite{xu2017online}, combine online learning with ensemble methods for their spatiotemporal forecast framework. This framework estimates the optimal weights for combining the ensemble member forecasts. Moreover, it uses an ``online'' algorithm that revises its previous forecasts when a future forecast is incorrect.

Yu \textit{et al.} \cite{yu2016hierarchical} propose a new approach to identify the hierarchical structure of spatio-temporal patterns at different resolution levels and subsequently construct a predictive model based on the identified structure. They first obtain indicators within different spatio-temporal spaces and construct distributed spatio-temporal patterns (DSTP). Next, they use a greedy searching and pruning algorithm to combine the DSTPs in order to form an ensemble spatio-temporal pattern (ESTP). The model, named CCRBoost, combines multiple layers of weighted ESTPs. They tested this method in predicting residential burglary, achieving 80\% accuracy on a non-publicly available dataset.

Recently, Wang \textit{et al.} \cite{wang2017deep} propose the use of deep learning for the prediction of hourly crime rates. In particular, their model ST-ResNet, extends the ResNet \cite{he2016deep} model for use in spatio-temporal problems like crime forecasting. They detected that future crime rates depend on the trend set in the previous week, the time of day and the nearby events both in space and in time. For each one of these contributing factors they use a separate ResNet model that offers a prediction based on indicators of a weekly period, a daily period and an hourly period respectively. The outputs of the 3 models are combined to form a common prediction. They also use external features like day of month, day of week and hour of day to get a more accurate prediction.

Wang \textit{et al.} propose a methodology very similar to Simonyan and Zisserman \cite{simonyan2014two}. Motivated by the fact that videos can be naturally decomposed into spatial and temporal components, Simonyan and Zisserman propose a two stream approach that breaks down the learning of video representation by using one stream to learn spatial and the other stream to learn temporal clues. For the spatial stream they adopt a typical CNN architecture using raw RGB images as input 
to detect appearance information. To account for temporal clues among adjacent frames, they explicitly generate multiple-frame dense optical flows derived from computing displacement vector fields between those frames.

Since this temporal stream operates on adjacent frames, it can only depict movements within a short time window. The second and most important problem of this method is that the order of the frames is not taken into account. Both problems are addressed by Wu \textit{et al.}\cite{wu2015modeling}, by replacing the temporal CNN stream with a Long-Short Term Memory (LSTM) \cite{hochreiter1997long} network, thus leveraging long term temporal dynamics. By fusing the outputs of the two streams, they jointly capture spatial and temporal features for video classification. They observed that CNNs and LSTMs are highly complementary. According to Ruta \textit{et al.}\cite{ruta2011generic} high complementarity of methods leads to ensemble methods that have excellent generalization ability.

\section{Problem Formulation} \label{problemFormulation}

Let $D$ be the dataset of $n$ four-dimensional vectors $\mathbf{x}_i$  with $i \in \{1,n\}$ where each $\mathbf{x}_i$ contains information on the location (longitude and latitude), the time and the type of a reported crime (crime category). 
It is typical in the literature to aggregate the spatial information by splitting the city map in a two-dimensional grid with cell edge size $l$, so that a $p$ cells square grid is produced (e.g. given a 8km by 8km sized map and a cell edge size $l=500m$, $p=16$ is calculated and thus a $16\times16$ cells grid is produced).

Next, the $\mathbf{x}_i$ data points are aggregated in each corresponding cell as a sum of occurrences of a certain crime type. Moreover, since the data are time-series of data points, the data are also split in time windows of duration $t$ and create multiple aggregated incident maps $I$ for the whole period $T$ of the time-series.

Our goal is to classify each cell as hotspot or not for a certain type of crime with the highest possible spatial resolution in a neighbourhood of the monitored city. Additionally, the hotspots that are predicted should be ranked according to the number of crimes that will occur inside their area for prioritizing and allocating policing resources more efficiently. In order to enhance the probability of an area being predicted as a hotspot, we use secondary parallel prediction of the number of occurrences $y$ for each crime.

Accordingly, the problem is defined as: a) a binary classification of the cells that will have occurrence of a certain crime in a certain time window in the future; b) the probability that a crime is classified as a hotspot is dependant on the number of occurrences $y$ of a certain crime for each cell in the defined future time window.

\section{Proposed Methodology} \label{methodology}


CNNs consist of convolutional layers, characterized by an input map $I$, a bank of filters $K$ and biases $b$, producing an output map $O$.
In the case of crime maps, by aggregating the crime incidents $\mathbf{x}_i$ to cells for every incident type and timespan $t$, we produce an ordered collection of incident maps $I$ for the whole duration $T$, with height $h$, width $w$ and $c$ channels such that $I \in \mathbb{R}^{h \times w \times c}$, analogous to a sequence of image frames from a video. Subsequently, for a bank of d filters with size $k_1$ and $k_2$, we have $K \in \mathbb{R}^{k_1 \times k_2 \times c \times d}$ and biases $b \in \mathbb{R}^{d}$, one for each filter. The output map $O \in \mathbb{R}^{h \times w \times d}$ is calculated by applying:
\begin{equation}
O_{ij} = \sum_{m=0}^{k_1-1} \sum_{n=0}^{k_2-1} \sum_{c=0}^{C} K_{m,n,c} \cdot I_{i+m, j+n, c} + b
\end{equation}
for every $i \in \{1,H-k_1\}$, $j \in \{1,W-k_2\}$, for each filter.

LSTMs, are a variant of RNNs that are better suited to long sequences since they do not suffer from the vanishing gradient effect \cite{hochreiter1997long}. An LSTM cell, maps the input vector $x^{(t)}\in \mathbb{R}^{n}$ for each timestep $t$ to the output vector $h^{(t)}\in \mathbb{R}^{m}$, by recursively computing the activations of the units in the network using the following equations:
\begin{equation}
\begin{aligned}
i^{(t)} &= \sigma (W_{xi}x^{(t)}+W_{hi}h^{(t-1)}+W_{hi}h^{(t)}+b_i) \\
f^{(t)} &= \sigma (W_{xf}x^{(t)}+W_{hf}h^{(t)}+W_{hf}h^{(t)}+b_f) \\
c^{(t)} &= f^{(t)}c^{(t-1)}+i_t \tanh(W_{xc}x^{(t)}+W_{hc}h^{(t)}+b_c)\\
o^{(t)} &= \sigma (W_{xo}x^{(t)}+W_{ho}h^{(t-1)}+W_{ho}h^{(t)}+b_o) \\
h^{(t)} &= o^{(t)}\tanh(c^{(t)})
\end{aligned}
\end{equation}
where $x^{(t)}$ and $h^{(t)}$ are the input and output vectors, $i^{(t)}$, $f^{(t)}$, $c^{(t)}$, $o^{(t)}$ are respectively the activation vectors of the input gate, forget gate, memory cell and output gate, and $W_{ab}$ denote the weight matrix from $a$ to $b$. In each time step $t$, the input of the LSTM cell consists of the input vector $x^{(t)}$ at time $t$ and the output vector $h^{(t-1)}$ from time step $t-1$.

Based on the methodology of Wu \textit{et al.} combination of CNNs and RNNs can be jointly used for a spatiotemporal forecasting model. Depending on the order that CNNs and LSTMs are used, 3 approaches present themselves. 

\begin{figure}[!t]
\centering
\subfloat[]{\includegraphics[width=0.47\textwidth]{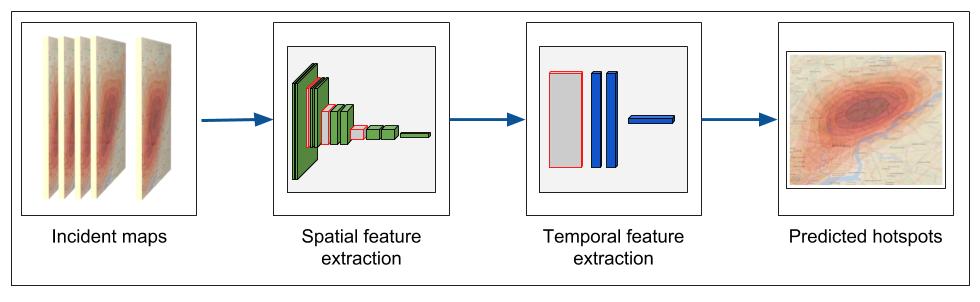}
\label{sftt}}
\hfil
\subfloat[]{\includegraphics[width=0.47\textwidth]{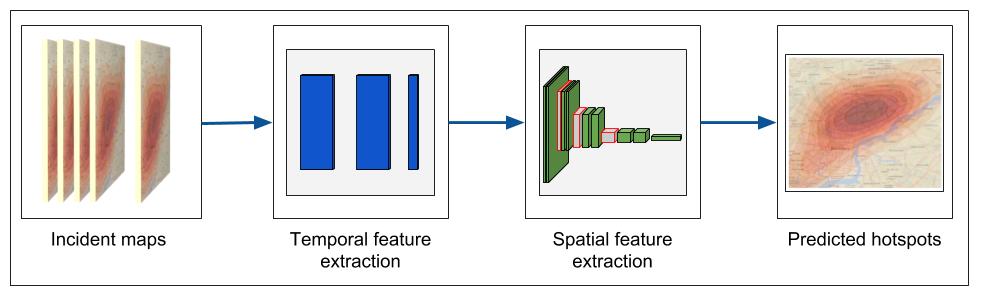}
\label{tfts}}
\hfil
\subfloat[]{\includegraphics[width=0.47\textwidth]{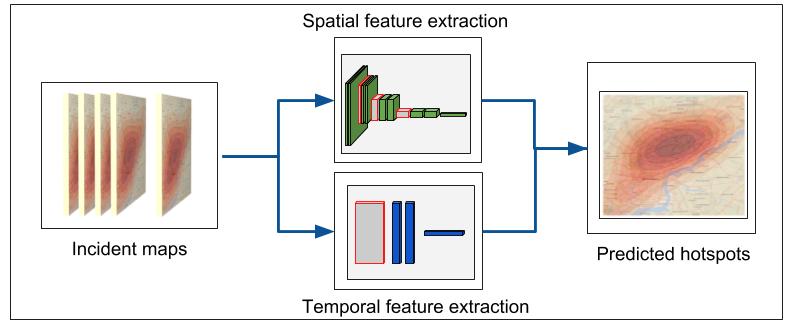}
\label{parb}}
\caption{Overview of the sequence of feature extraction: (a) the spatial features first then the temporal (SFTT), (b) the temporal features first then the spatial (TFTS) and (c) spatial and temporal features in two parallel branches (ParB)}
\label{arch_overview}
\end{figure}

The first approach, named SFTT (Fig. \ref{sftt}), passes each incident map from a CNN submodel, producing a feature vector for every timespan $t$ that encodes the spatial distribution of incidents in a feature space that is much smaller than the original. The sequence of $\frac{T}{t}$ feature vectors is then fed into the LSTM network which can extract temporal features.
All 11 crime categories are used as input for the incident maps in separate input channels. The intuition is that the ratio of crime types in each area encodes implicitly the socioeconomic status and activities’ profile of the area and thus provides better modeling of the situation. The justification is presented in Section \ref{sec:res:eval_spatialBody}.

The second approach is to input the feature maps to an RNN submodel for temporal feature extraction and then use a CNN for spatial features. For this approach, named TFTS (Fig. \ref{tfts}), we firstly extract a temporal feature vector for each cell by passing each sequence of incidents $\mathbf{x}_i$ of each cell through an LSTM network model. The extracted temporal features for each cell retain the relative position on the monitored area (i.e. their cell position), thus temporal maps of the area are created. The temporal maps are then used as input features that are fed to a CNN network for spatial information extraction.

The third possibility is to extract spatial and temporal features in parallel, named ParB (Fig. \ref{parb}), using 2 separate branches. The output of the 2 branches can be combined so that prediction accounts for both groups of features. 

\begin{figure}[!t]
\centering
\subfloat[]{ 
\includegraphics[width=3.3in]{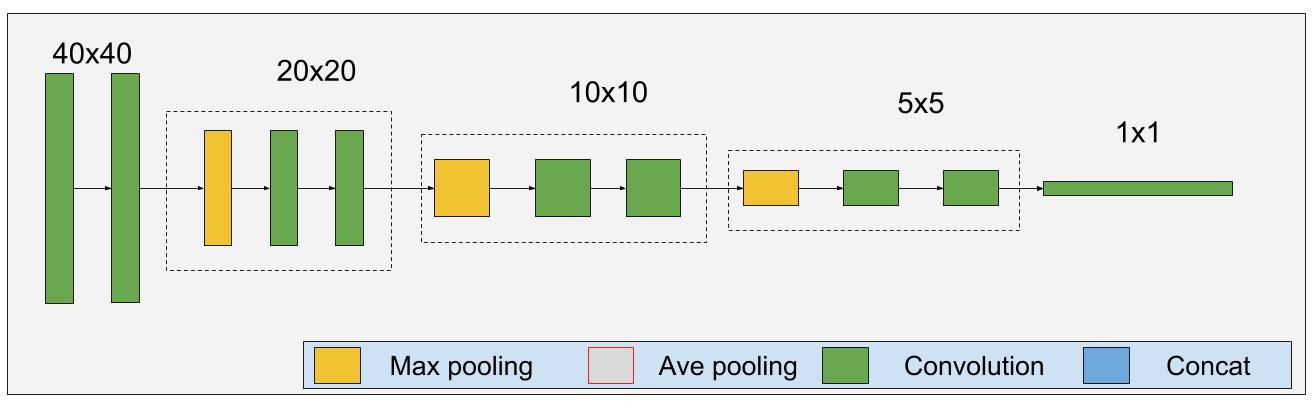}
\label{vgg_graph}
}\hfil
\subfloat[]{ 
\includegraphics[width=3.3in]{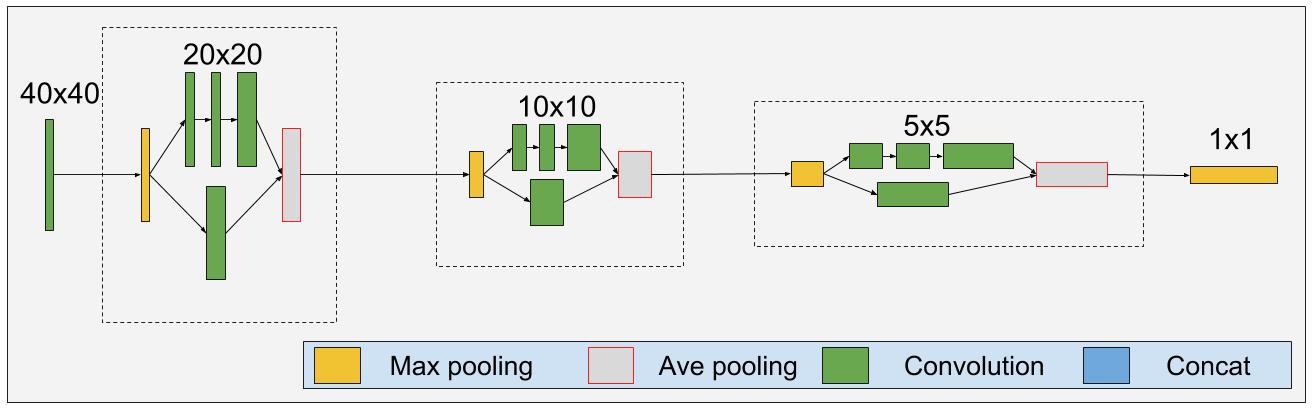}
\label{resnet_graph}
}\hfil
\subfloat[]{ 
\includegraphics[width=3.3in]{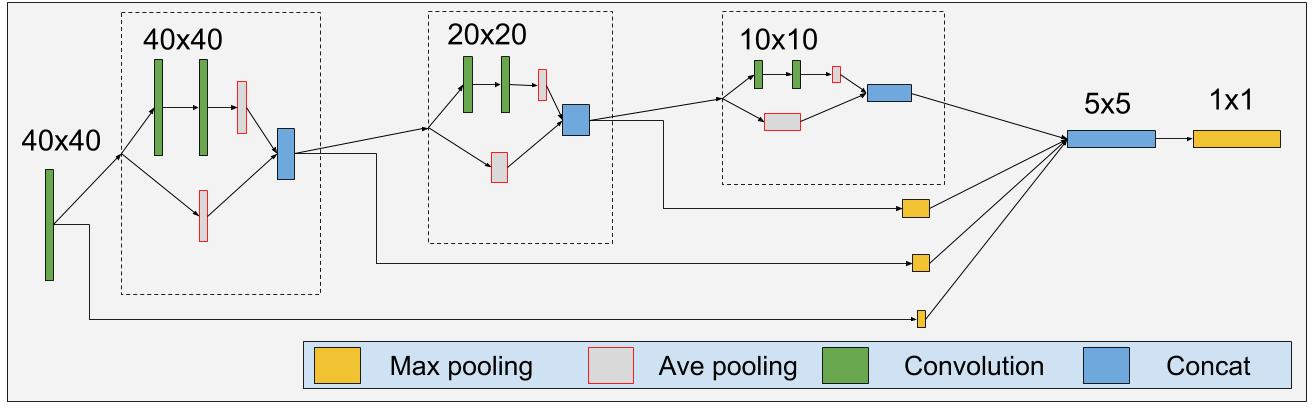}
\label{fastmask_graph}
}\hfil
\subfloat[]{ 
\includegraphics[width=3.3in]{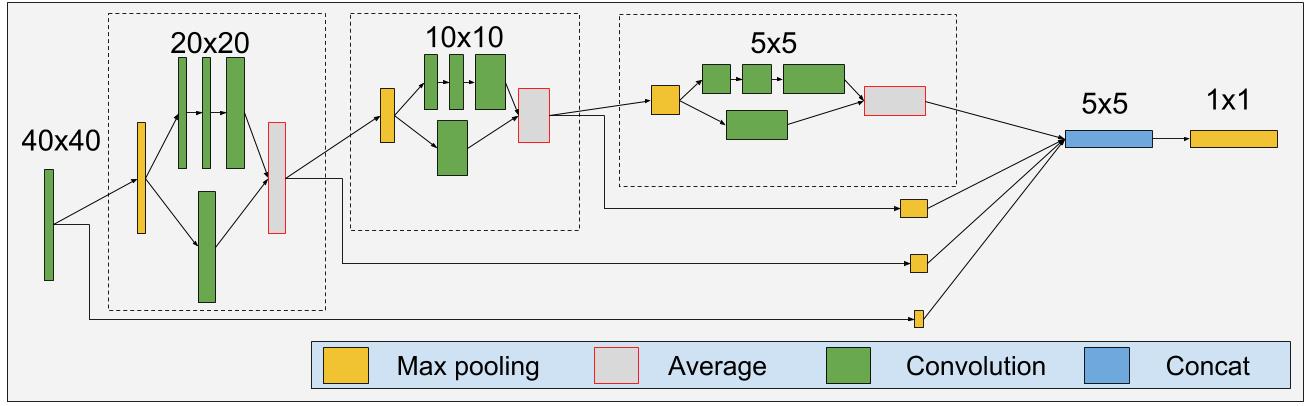}
\label{fastresmask_graph}
}\hfil
\caption{Convolutional network body architectures used for the extraction of spatial information: (a) VGGNet, (b) ResNet, (c) FastMask, (d) FastResidualMask. Convolutional layers are coloured green, pooling layers are coloured yellow and the grey are non-parametric layers (i.e. concatenation)}
\label{bodies}
\end{figure}

For the spatial feature extraction, we explore four possible body architectures depicted in Fig. \ref{bodies}. Three out of them are based on VGGNet \cite{simonyan2014very}, ResNet \cite{he2016deep} and FastMask \cite{hu2016fastmask}, respectively, while the fourth one is a combination of the ResNet and the FastMask models. 
All these models were originally designed to extract features from images of size $224\times224$ pixels. In order to follow the grid resolution ranges used in related crime prediction papers \cite{yu2016hierarchical, eck2005mapping, chainey2008utility, chainey2013gis},
the input grid resolution had to be reduced. Thus, similar models but with smaller input and fewer parameters were implemented.

A common practice in DL is to take models that are trained in one dataset and use them in another dataset, sometimes even in different domains. One reason that we did not use the ImageNet \cite{ImageNet} pre-trained models of VGG, ResNet and FastMask is that we altered the models themselves. Another reason is that the ImageNet dataset is not only of a different domain but the structural and statistical characteristics of images is completely different than incident maps that a random start is preferred.

In our VGGNet body, we use series of 5 pairs of convolution layers, where the first 3 are followed by pooling layers. In the first 4 pairs of convolution layers the filter size we use is $3\times3$, while in the last pair the filter size is $1\times1$. The first 2 pairs of convolutions apply 32 filters, the following 2 pairs apply 64 filters and the last pair applies 256 filters. In all convolution function layers the activation function is the Rectified Linear Units (ReLU), which allows the networks to converge faster \cite{simonyan2014very}. The pooling layers use the max operation. We use a Batch Normalization layer and a Dropout layer after each pooling layer, since our datasets are fairly small and the possibility of overfitting is significant. The complete CNN body of this network is depicted in Fig. \ref{vgg_graph}.

A more complex CNN that has proven to be more effective in extracting spatial information in image classification than VGGNet is ResNet \cite{he2016deep}. In ResNet, the added residual layers aim to solve the degradation problem that was encountered when the convolutional networks became too deep. In every convolution block there is a shortcut connection added to each pair of $3\times3$ filters. Inspired by this tactic, we modified the feature extractor to incorporate residual connections as shown in Fig. \ref{resnet_graph}. In each block, we use 3 convolution layers with $3\times3$, $1\times1$ and $3\times3$ sized filters and a parallel residual convolution layer with a filter size of $3\times3$. In each consecutive block, the number of filters in the last convolution layer and the residual layer, is doubled with regard to the previous block. The number of filters in the first 2 convolution layers is a quarter of the last. The output of the block layers with the output of the residual layer is then averaged before a max pooling layer changes the scale.

In FastMask \cite{hu2016fastmask}, Hu \textit{et al.} use a block of convolutional layers (called a neck) in order to extract features of different scales. The features are then concatenated before they are forwarded to the next neck. The output of all the necks is also concatenated before passed on to the classification layers. In order to change scale, in each neck there is an average pooling layer. Each neck has 2 paths for information to flow. In the one path, the input of the neck is passed through an average pooling layer, thus zooming out without extracting any new features. The other path uses 2 convolutional layers with $3\times3$ sized filters, extracting this way a number of features from this scale. In Fig. \ref{fastmask_graph} we illustrate our own implementation of the FastMask model.

The two methods differ in their philosophy of using the residual information. In ResNet the blocks are used sequentially, while in FastMask the outputs of all blocks are brought together in one last concatenation layer. Moreover, the residual information is averaged with the new features in the ResNet design, while in FastMask the residual information is concatenated to the newly extracted features. By using the blocks of ResNet in the place of FastMask necks, we create a variation where averaged main and residual features from each scale are then concatenated (Fig. \ref{fastresmask_graph}).

The temporal feature extraction in the first approach is achieved by extracting the temporal correlations in the sequence of vectors holding the spatial features by using an LSTM submodel. This LSTM model consists of 3 consequtive LSTM layers where the 2 first ones have 500 neurons each, while the last one has the same number of neurons as the number of cells that the model predicts. In the second and third approaches, an LSTM submodel with 3 layers of 32 neurons is applied to each cell of the input. Each LSTM layer adds another level of non-linearity in the extraction of temporal patterns from the input data.

In all proposed models, the extracted spatio-temporal features are finally fed into two parallel fully connected output layers. On the one output layer we use the binary cross entropy (BCE) loss function in order to classify each cell as a hotspot or not:
\begin{equation}
BCE = -\frac{1}{N} \sum_{i}^{N}[\hat{o}_i \log{o_i} + (1-\hat{o}_i) \log{(1-o_i)}]
\end{equation}
where $N$ is the total number of cells, $\hat{o}_i$ is the predicted class of cell $i$ and $o_i$ is the actual class of cell $i$.

On the other output layer we apply the mean squared error (MSE) loss function:
\begin{equation} \label{eq:mse}
MSE = -\frac{1}{N} \sum_{i}^{N} (y_i - \hat{y}_i)^2
\end{equation}
where $N$ is the total number of cells, $\hat{y}_i$ is the predicted number of crimes that will occur in the cell during the target time period and $y_i$ is the actual number of crimes that occurred inside cell $i$ during the target time period.

The loss from both outputs is combined during the back propagation step, so that the spatio-temporal features that are extracted, affected by the classification of a cell as a hotspot and to the number of crimes that occurred within the cell. This approach was selected to drive the final classifier to give bigger probabilities to cells where multiple crimes occur.


By changing the binary cross entropy loss function to the equivalent multi-class cross entropy (equation \ref{eq:mcce}), we are able to predict the hotspot distribution for all the crimes at the same network.
\begin{equation} \label{eq:mcce}
MCCE = -\frac{1}{N} \sum_{i}^{N} \sum_{k}^{C}[\hat{o}_{i k} \log{o_{i k}}]
\end{equation}

\section{Experimental Setup} \label{experimentalSetup}

\subsection{Algorithms}
In order to have a proper assessment of the capability of the DL methods, we compare them with the state-of-the-art CCRBoost \cite{yu2016hierarchical} and ST-ResNet \cite{wang2017deep} methods, as well as eight baseline methodologies that commonly appear in the recent relevant literature.

\begin{enumerate}

\item CCRBoost \cite{yu2016hierarchical}. CCRBoost starts with multi-clustering followed by local feature learning processes to discover all possible distributed patterns from distributions of different shapes, sizes, and time periods. The final classification label is produced using groupings of the most suitable distributed patterns. 

\item ST-ResNet \cite{wang2017deep}. The original ST-ResNet model uses 3 submodels with residual connections, that each has 4 input channels in parallel, in order to extract indicators from 3 trends: previous week, time of day and recent events.
 In our problem, the temporal resolution is not hourly but daily so the 3 periods are replaced by day of month, day of week and recent events equivalently.

\item Decision Trees(C4.5) \cite{quinlan2014c4} using confidence factor of 0.25; Decision Trees is a non-parametric supervised learning method that predicts the value of a target variable by learning simple decision rules inferred from the data features.

\item Naive Bayes \cite{zhang2004optimality} classifier with a polynomial kernel; Naive Bayes methods are a set of supervised learning algorithms based on applying Bayes' theorem with the “naive” assumption of independence between every pair of features.

\item LogitBoost \cite{friedman2000additive} using 100 as its weight threshold; The LogitBoost algorithm uses Newton steps for fitting an additive symmetric logistic model by maximum likelihood.

\item Random Forests \cite{breiman2001random} with 10 trees; A random forest is a meta estimator that fits a number of decision tree classifiers on various sub-samples of the dataset and use averaging to improve the predictive accuracy and control over-fitting.

\item Support Vector Machine (SVM) \cite{cortes1995support} with a linear kernel; SVMs are learning machines implementing the structural risk minimization inductive principle to obtain good generalization on a limited number of learning patterns.

\item k Nearest Neighbours \cite{arya1998optimal} with 3 neighbours; kNN is a classifier that makes a prediction based on the majority vote of the k nearest samples on the feature vector space.

\item MultiLayer Perceptron (MLP(150)) \cite{olligschlaeger1997artificial} with one hidden layer of 150 neurons; 

\item MultiLayer Perceptron (MLP(150,300,150,50)) \cite{ruck1990multilayer} with four hidden layers of 150, 300, 150 and 50 neurons each; 

\end{enumerate}

%
%
%
%
%
%
%
%
%

The CCRBoost algorithm was reimplemented by us in python, 
ST-ResNet is based on DeepST \cite{zhang2017deep} which we downloaded from github\footnote{https://github.com/lucktroy/DeepST/tree/master/scripts\\/papers/AAAI17} and adapted,
while for the rest of the baseline methods we used the implementations available from scikit-learn \cite{pedregosa2011scikit}.
Our DL experiments were implemented in the Keras framework \cite{chollet2015keras} using the tensorflow \cite{tensorflow2015whitepaper} backend.


The complexity of the algorithms is compared, in terms of computational time and the number of learnable parameters, in Table \ref{tbl:complexity}. The baseline algorithm are typically much faster, however the DL based ones are also equally applicable since they need minutes of computation to return results. The only exception is SFTT-FastResmask which required approx. 2 hours to converge.
All experiments were performed on a 12core 3.3GHz linux system with 64GB RAM and an Nvidia TitanX GPU with CUDA \cite{nvidia2010programming}.

\begin{table}[t]
\small
\caption{Complexity of all algorithms in total training time for the ``All Crimes'' crime type in the Philadelphia dataset and number of trainable parameters for the DL architectures}
\label{tbl:complexity}
\resizebox{\columnwidth}{!}{%
\begin{tabular}{ l c c }
\hline
Algorithm & Training time & \# Parameters \\
\hline
CCRBoost & 0:00:38 & - \\
Decision Trees (C4.5) & 0:00:02 & - \\
Naive Bayes & 0:00:03 & - \\
Logit Boost & 0:00:05 & - \\
SVM & 0:01:17 & - \\
Random Forests & 0:00:03 & - \\
KNN & 0:00:02 & - \\
MLP (150) & 0:00:13 & - \\
MLP (150, 300, 150, 50) & 0:00:24 & - \\
ST-ResNet & 0:05:59 & 1.343.043 \\
\hline
SFTT-VGG19 & 0:48:11 & 30.117.120  \\
TFTS & 0:24:32 & 10.260.016  \\
ParB & 3:15:05 & 31.942.280  \\
SFTT-ResNet & 0:17:53 & 7.348.899  \\
SFTT-FastMask & 0:17:57 & 6.917.264  \\
SFTT-FastResMask & 1:53:09 & 7.610.299  \\
\hline
\end{tabular}
}
\end{table}

\subsection{Datasets}

In the last couple of years a number of new datasets have been published in the field of crime prediction, mainly from law enforcement agencies in the US. In this paper, we have selected and used in our experiments, 5 of the most prominent open datasets that can be downloaded from Kaggle\footnote{https://www.kaggle.com/}. These 5 datasets include incident reports from Seattle\footnote{https://www.kaggle.com/samharris/seattle-crime}, Minneapolis\footnote{https://www.kaggle.com/mrisdal/minneapolis-incidents-crime}, Philadelphia\footnote{https://www.kaggle.com/mchirico/philadelphiacrimedata}, San Fransisco\footnote{https://www.kaggle.com/c/sf-crimedatasets/} and Metropolitan DC\footnote{https://www.kaggle.com/vinchinzu/dc-metro-crime-data} police departments. While each dataset contains a number of unique attributes, all 5 datasets report a location (in latitude and longitude), a time and a type of event for each incident $\mathbf{x}_i$. 
The basic information for each dataset is presented in Table \ref{tbl:dataset}.

\begin{table}
\caption{Basic statistics for the 5 examined datasets}
\label{tbl:dataset}
\begin{tabular}{ l c c r }
\hline
Dataset & start year & end year & Num. of incidents  \\
\hline
Philadelphia & 2006 & 2017 & 2,203,785 \\
Seattle & 1996 & 2016 & 684,472  \\
Minneapolis & 2010 & 2016 & 136,121 \\
DC Metro & 2008 & 2017 & 313,410 \\
San Francisco & 2003 & 2015 & 878,049 \\
\hline
\end{tabular}
\end{table}

\begin{figure}[!t]
\centering
\includegraphics[width=3.4in]{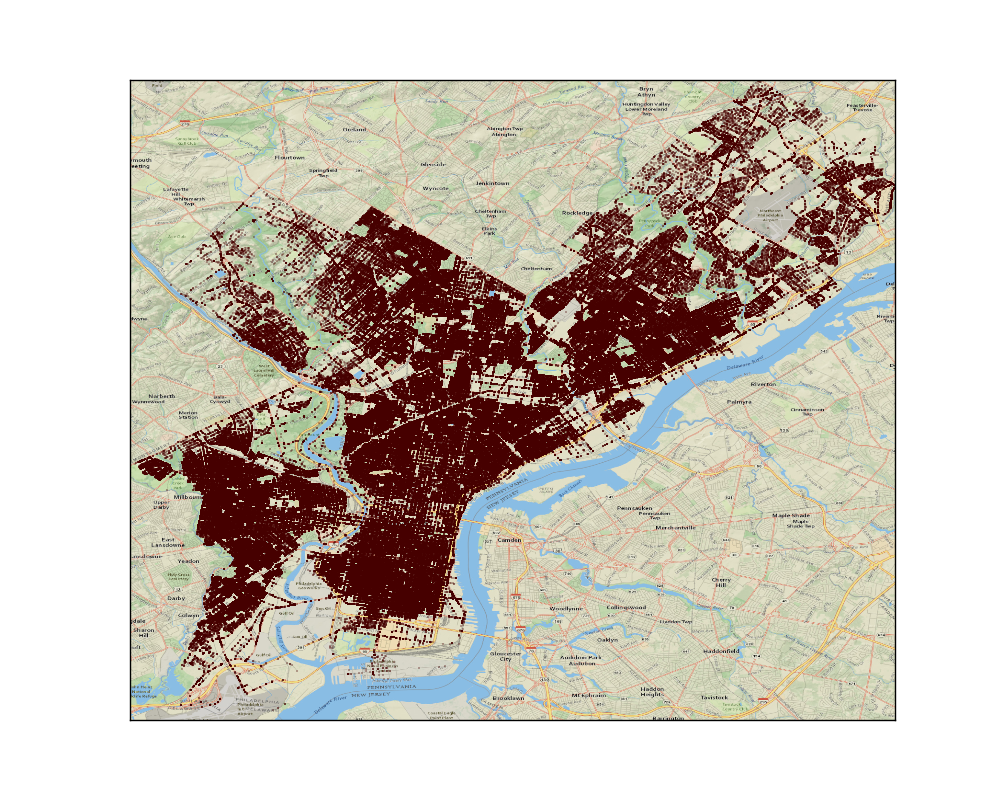}
\caption{Philadelphia map overlayed with a plotting of all incidents that occured in 2006.}
\label{philly}
\end{figure}

An example plotting of all incidents from 2006 in the Philadelphia dataset is depicted in Fig. \ref{philly}. 

Some of the datasets have multiple time attributes recorded, for example, the time when an incident took place and the time when it was reported. Since the report time is recorded by automatic systems, we selected to use this time feature when available. 

The datasets include different codes for event types, different levels of description for event types and different event types that are recorded. In order to mitigate the discrepancies as much as possible, we decided to homogenize the provided data classes into 10 crime types (i.e. ``Homicide'', ``Robbery'', ``Arson'', ``Vice'', ``Motor Vehicle'', ``Narcotics'', ``Assault'', ``Theft'', ``Burglary'',``Other'') and assign each of the datasets' categories to one of these. We defined the ``Other'' category for the crime types that do not easily fall into the other categories or where the location is highly constrained or totally irrelevant to the crime type as is for example fraud and embezzlement. Moreover, we used one class for all aforementioned types of crime, which aims to encapsulate high crime areas irrespective of the crime type, named ``All Crimes''. A detailed presentation of the classes of each dataset and the homogenization approach we followed, is reported in 
Appendix A. 

The spatial resolution of all datasets is enough for a block-level analysis of crime. Being consistent with the grid resolution ranges that are used in related crime prediction papers 
\cite{yu2016hierarchical, eck2005mapping, chainey2008utility, chainey2013gis} 
the finer block size resolution is defined to have cell edge size $l \approx 450m$, resulting in grids of $40\times40$ cells. Using an even finer resolution, the resulting grids would be extremely sparse, especially in crime types with a small number of incidents. The coarser resolution of neighbourhoods has cells with approximate cell edge size $l \approx 800m$, leading to grids of $16\times16$ cells. For the different datasets, instead of adjusting the number of cells according to actual distances in the cities, we opted to use a fixed number of cells overlayed on the area under investigation for each dataset.

We aim to create a setting where emerging hotspots are predicted in advance by evaluating past incidents in the current month. 
Thus, the past incidents are aggregated in incident maps $I$ of timespan $t$ of 1 day, and for a period $T$ of 30 days i.e. 30 daily incident maps are used as input to forecast crimes for the next period. We used a daily timespan to aggregate incidents $\mathbf{x}_i$ so that enough temporal detail can be extracted while the time series are sufficiently populated.

The smallest of our datasets describes just over 4 years of data. For this reason, we extracted from all the datasets 3 years of incidents to use as training data and 1 year of incidents for testing purposes. Testing is performed by making a prediction for every month and the reported scores are the mean of the twelve individual prediction scores.
The number of samples is relatively small but is enough to evaluate which of the proposed architectures can perform better.

Each cell is marked as a hotspot, if at least one incident occurred in the following month, otherwise a coldspot.
After the crime categories are merged, for the training time period, the average number of hotspots per day is presented in Table \ref{tbl:meancells} for every crime type. The cells that had no activity in the total duration of the experiments are considered outside the study area and were removed from the metrics calculation. In Table \ref{tbl:meancells} we only present the number of cells that remain inside the study area.
From these numbers we observe that especially in the sparsest crime types there is a class imbalance problem. In order to avoid it, we have selected metrics that take this fact into account.

\begin{table}
\caption{Mean number of hotspots per day for every crime type and dataset for the highest resolution of 40 cells by 40 cells}
\label{tbl:meancells}
\resizebox{\columnwidth}{!}{%
\begin{tabular}{ l c c c c c }
\hline
Crime Type & Philadelphia & Seattle & Minneapolis & DC Metro & San Francisco \\
\hline
ASSAULT & 80.28 & 12.84 & 3.08 & 6.32 & 14.96 \\
THEFT & 116.72 & 21.34 & 17.46 & 27.39 & 40.44 \\
ROBBERY & 20.74 & 2.99 & 6.13 & 10.56 & 4.74 \\
BURGLARY & 23.76 & 17.06 & 12.36 & 9.81 & 8.65 \\
MOTOR VEHICLE & 27.94 & 10.01 & 14.44 & 8.43 & 7.77 \\
ARSON & 1.37 & 0.10 & 0.32 & 0.10 & 0.29 \\
HOMICIDE & 0.80 & 0.03 & 0.74 & 0.27 & 0.0 \\
VICE & 4.36 & 22.59 & 0.72 & 0.59 & 1.59 \\
NARCOTICS & 27.90 & 3.21 & 0.0 & 0.0 & 5.54 \\
OTHER & 102.37 & 40.15 & 0.09 & 23.24 & 51.81 \\
\hline
\# CELLS & 752 & 818 & 1123 & 702 & 1057 \\
\hline
\end{tabular}
}
\end{table}

\subsection{Metrics}


When dealing with data that have class imbalance, the metric of accuracy is not very informative because by always predicting the most dominant class the scores will be very high. $F1score$ is preferred because it is the harmonic mean of precision and recall. Precision is the number of correct positive results divided by the number of all positive results, and recall is the number of correct positive results divided by the number of positive results that should have been returned. The $F1score$ is calculated by:
\begin{equation}
F1score = 2 * \frac{precision * recall} {precision + recall}
\end{equation}
With the $F1score$ we measure the ability of the methods to both correctly predict hotspots and how many of the hotspots we identified at the same time.

$AUROC$ (Area Under Curve - Receiver Operating Characteristic) is the calculated area under a receiver operating characteristic curve. The ROC curve plots parametrically $TPR(m)$ versus $FPR(m)$ with $m$ being the probability threshold used to classify a prediction as hotspot.
The $TPR$ (True Positive Rate) and $FPR$ (False Positive Rate) are defined as:
\begin{equation}
TPR = \frac{True_{hot}}{True_{hot} + False_{cold}}
\end{equation}
\begin{equation} 
FPR = \frac{False_{hot}}{False_{hot} + True_{cold}}
\end{equation}
The area under the curve summarizes the performance of a classifier for all possible thresholds and is equal to the probability that a classifier will rank a randomly chosen hotspot higher than a randomly chosen coldspot, when using normalized units \cite{ROC}.
\begin{equation}
AUROC = \int_{\infty}^{-\infty} {TPR}(m) - {FPR}^\prime(m) dm
\end{equation}

$AUCPR$ (Area Under Curve - Precision Recall) equivalently, is the calculated area under a precision-recall curve. A precision recall curve plots parametrically $precision(m)$ versus $recall(m)$ with $m$ the varying parameter as above, where:
\begin{equation}
precision = \frac{True_{hot}}{True_{hot} + False_{hot}}
\end{equation}
\begin{equation}
recall = \frac{True_{hot}}{True_{hot} + False_{cold}}
\end{equation}

\textit{PAI} (Prediction Accuracy Index), which is defined by Chainey \textit{et al.}\cite{chainey2008utility} specifically for crime prediction and measures the effectiveness of the forecasts with the following equation:
\begin{equation}
PAI = \frac{\frac{r}{R}} {\frac{a}{A}} 
\end{equation}
where $r$ is the number of crimes that occur in an examined forecast area, $R$ is the total number of crimes in the entire map, $a$ is the forecast area size, and $A$ is the area size of the entire map under study.
The PAI metric offers a useful insight to the effectiveness of the forecast each method produces, since it not only measures the number of correct hotspots but takes into account the importance of each hotspot, which is our ultimate goal. 
This metric depends heavily on the percentage of the total area that is predicted to be hot. In order to simulate pragmatic conditions where a law enforcement agency has a limit to the available resources for fighting crime, we limit the maximum area that can be predicted as hot to 5\% of the total area and report this metric as PAI@5.


\section{Results} \label{results}
The selected experimental approach was to evaluate which of the proposed DL methods is the most robust compared with the state-of-the-art algorithms and select it for further analysis. In Section \ref{sec:res:eval_models} the comparison against all algorithms, datasets and metrics is presented. Section \ref{sec:res:eval_cellsize} examines the robustness of the models in varying cell sizes. Sections \ref{sec:res:eval_spatialBody} and \ref{sec:res:eval_tempBody} present different submodel configurations on the selected DL approach. Section \ref{sec:res:eval_batchNormDropout} presents the effect of batch normalization and dropbout techniques. Finally, the impact of multi-label classification compared with the binary approach is examined in Section \ref{sec:res:eval_multi}. 

\subsection{Evaluation of Models}
\label{sec:res:eval_models}
\begin{figure}[!t]
\centering
\subfloat[]{
	\includegraphics[width=2.6in, height=1.6in]{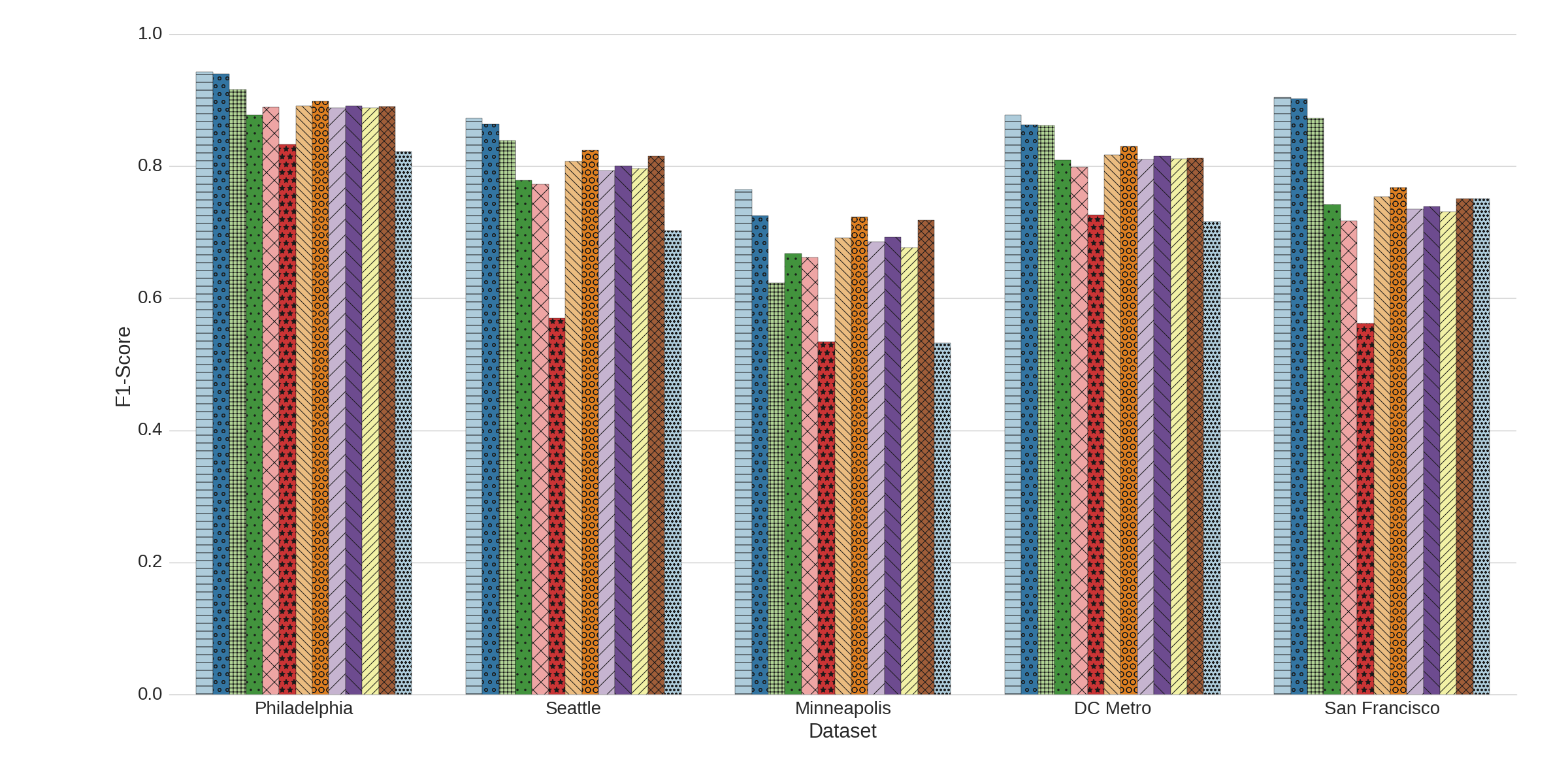}
	\label{per_dataset_f1score}
}
\hfill
\subfloat[]{
	\includegraphics[width=2.6in, height=1.6in]{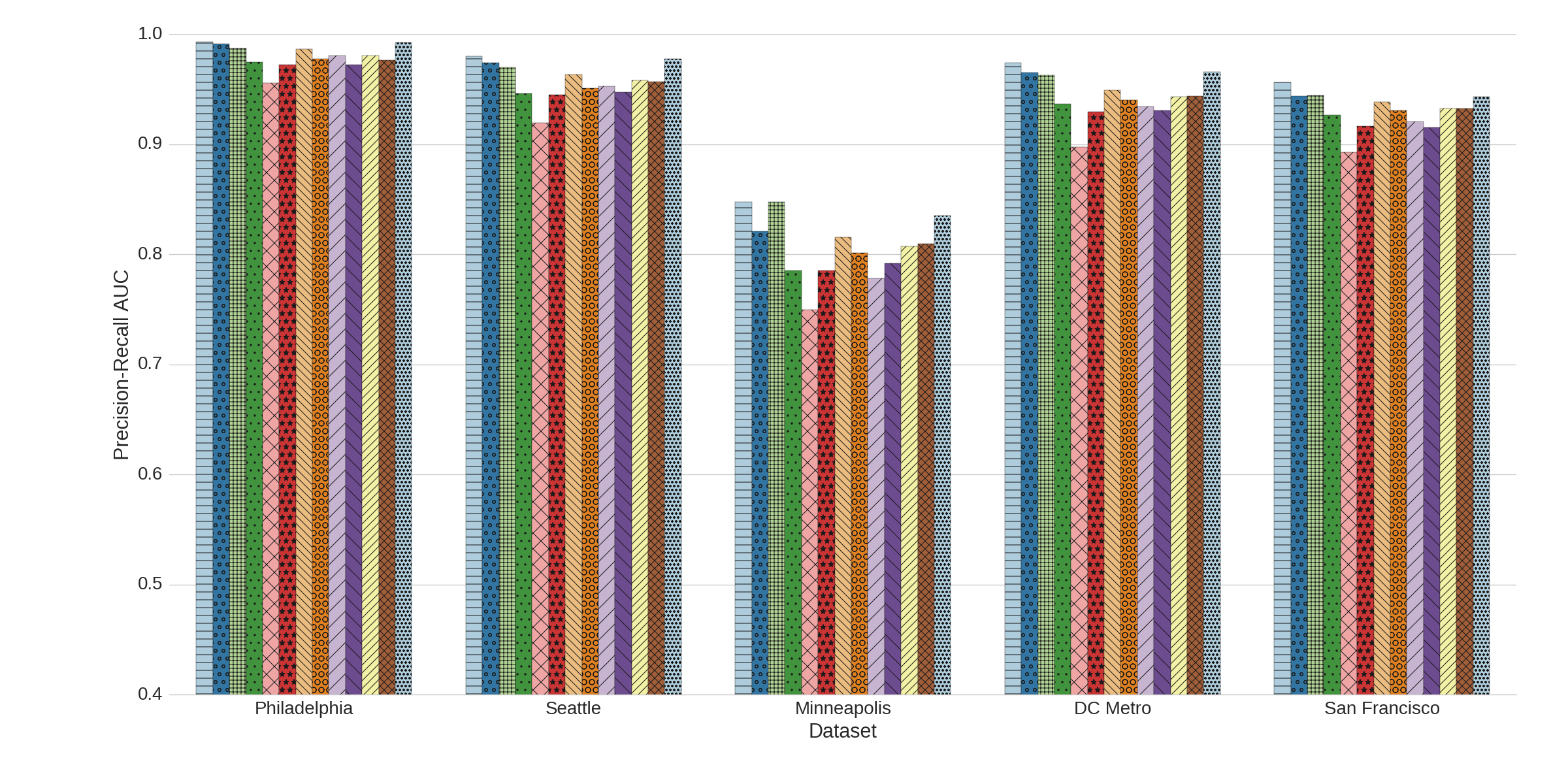}
	\label{per_dataset_pr_auc}
}
\hfill
\subfloat[]{
	\includegraphics[width=2.6in, height=1.6in]{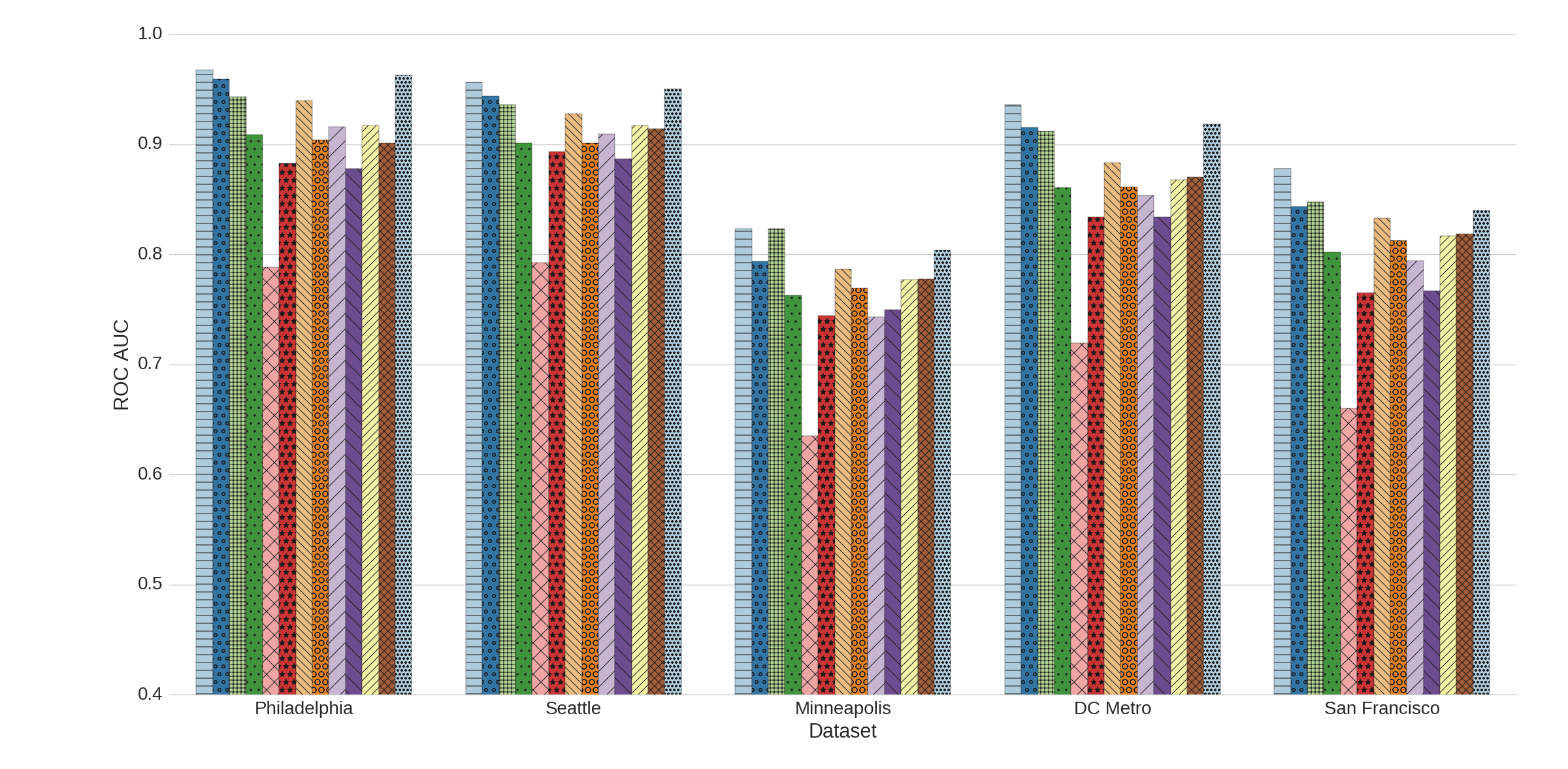}
	\label{per_dataset_roc_auc}
}
\hfill
\subfloat[]{
	\includegraphics[width=2.6in, height=1.6in]{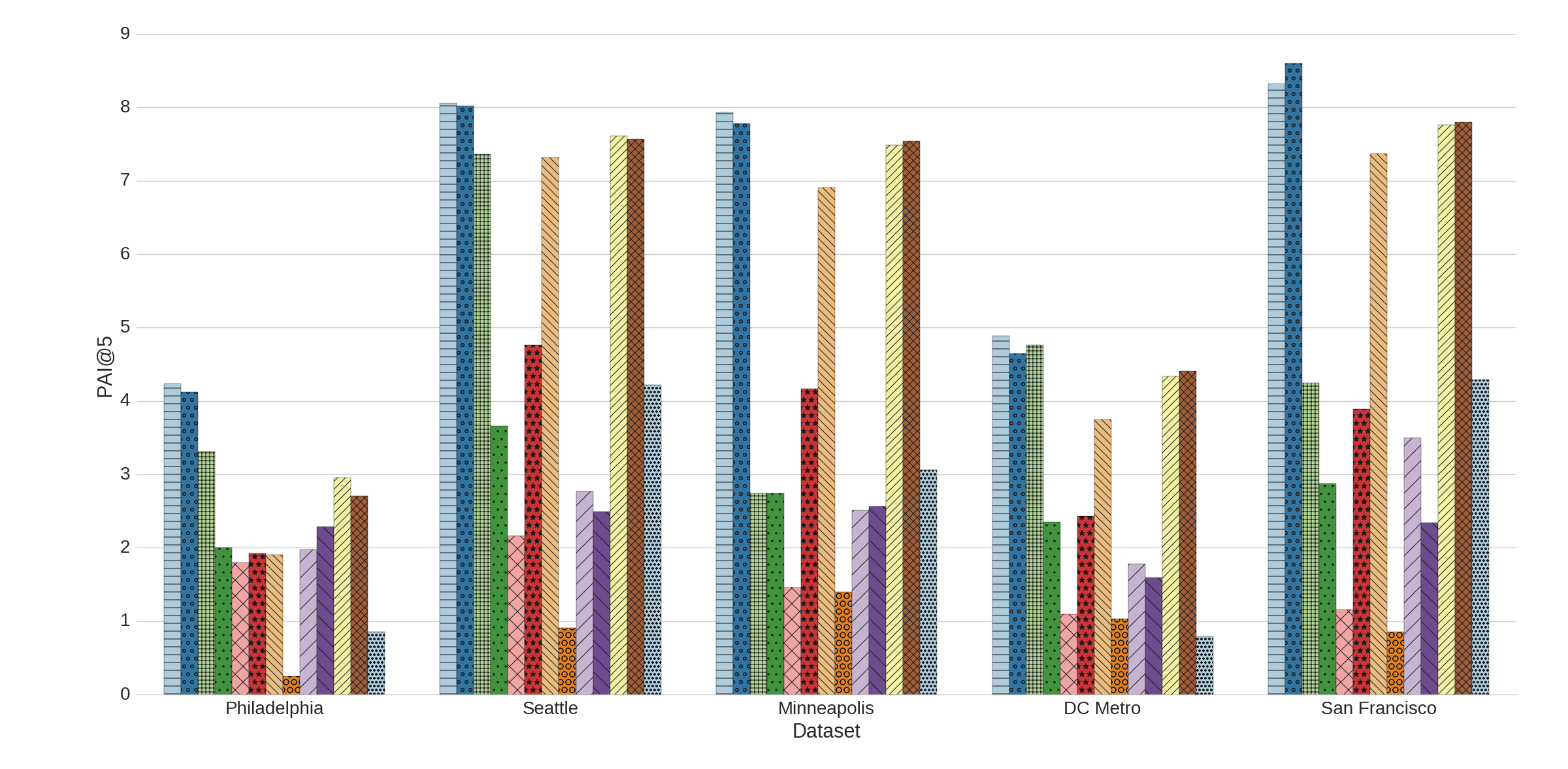}
	\label{per_dataset_pai5}
}
\hfill
\subfloat{
	\includegraphics[width=3.34in]{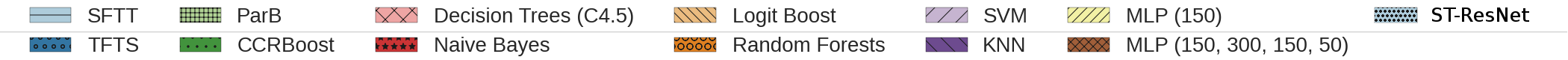}
}
\hfill
\caption{(a) F1score, (b) PR AUC and (c) ROC AUC per dataset for ``All Crimes'' crime type for the 3 DL models and 10 baseline approaches for cell size of 450m (40 by 40). Best viewed in color.}
\label{per_dataset}
\end{figure}

\begin{figure}[!t]
\centering
\subfloat[]{
	\includegraphics[width=3.0in, height=1.55in]{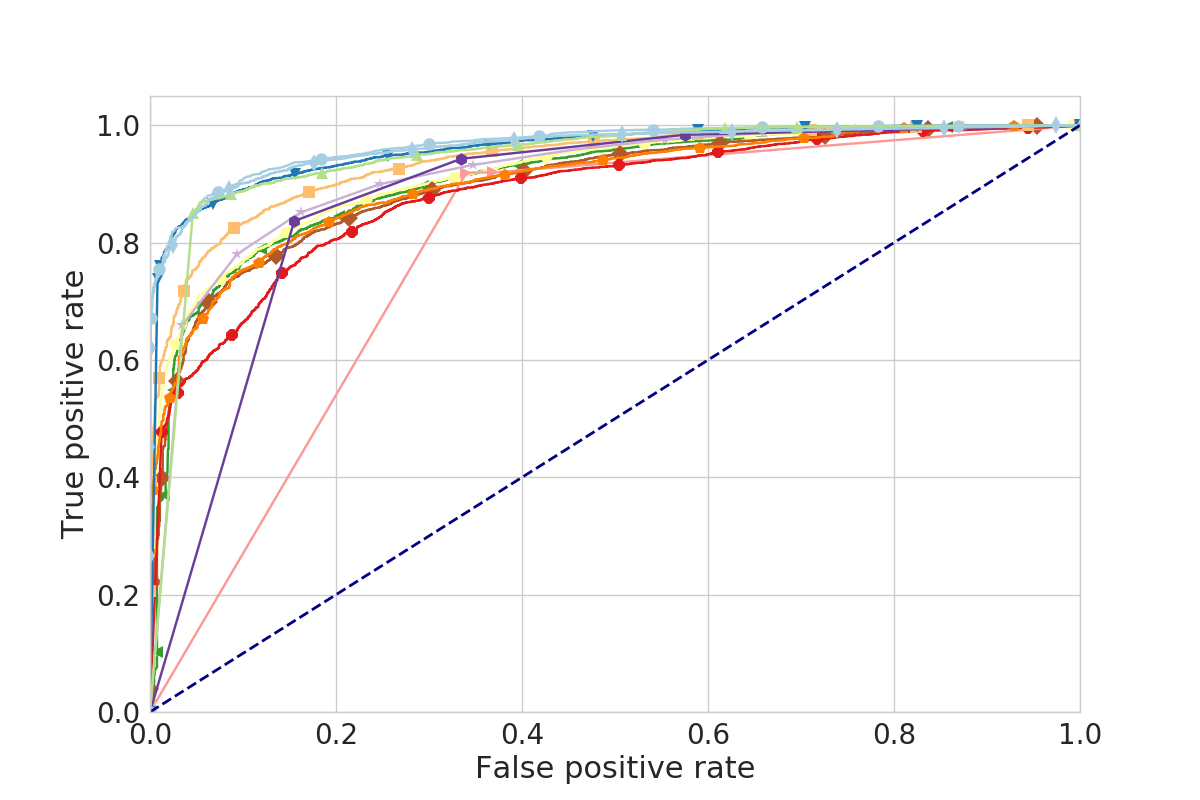}
	\label{roc_curves}
}\hfill
\subfloat[]{
	\includegraphics[width=3.0in, height=1.55in]{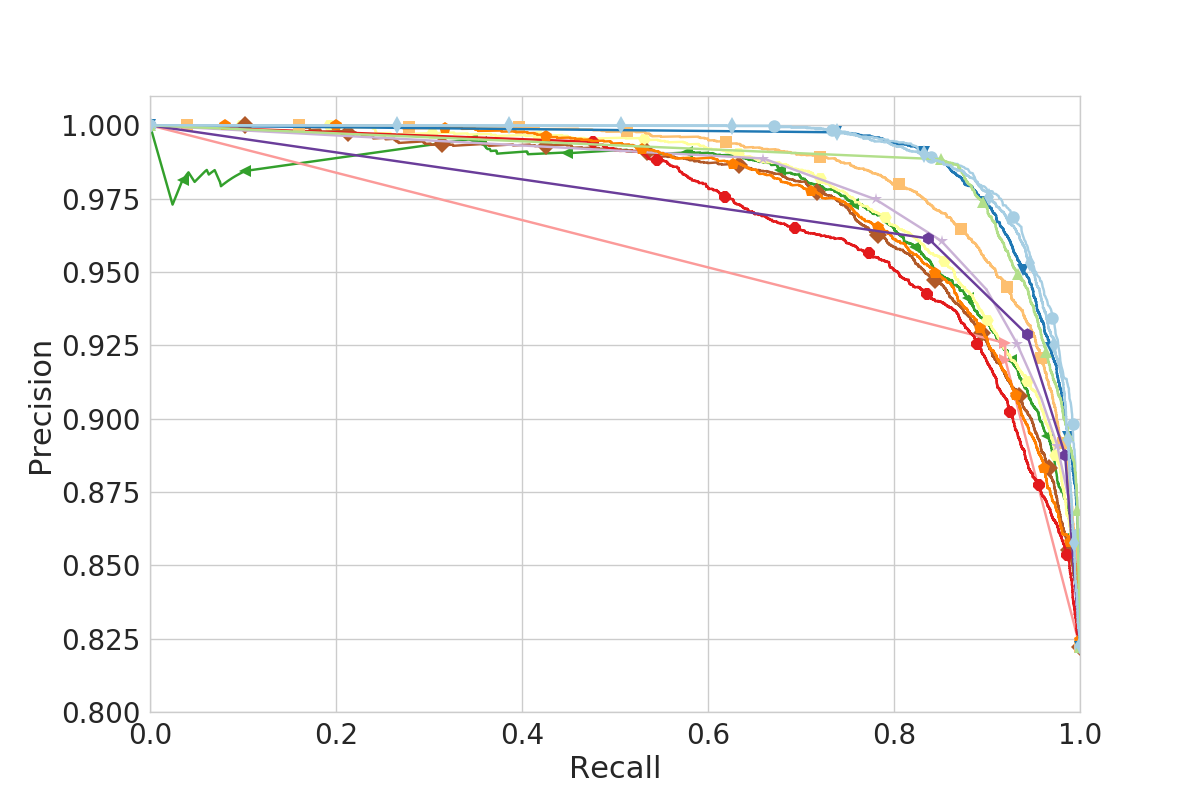}
	\label{pr_curves}
}\hfill
\subfloat{
	\includegraphics[width=3.3in]{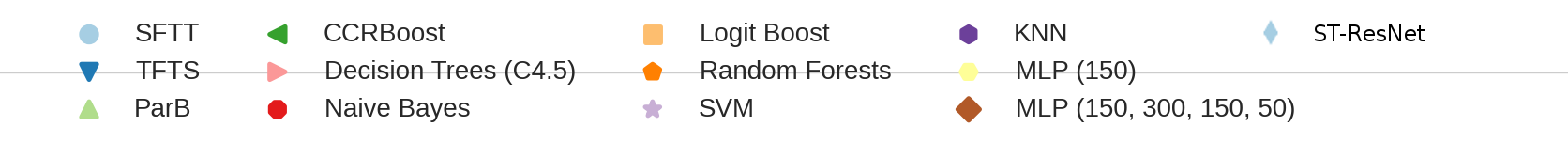}
}
\hfill
\caption{(a) ROC Curves and (b) Precision-Recall Curves for ``All Crimes'' crime type for the 4 DL bodies and 10 baseline approaches in the Philadelphia dataset. Best viewed in color.}
\label{curves_per_crime}
\end{figure}

By comparing the three main model building approaches, we investigate if the data are more correlated in the spatial or temporal axes. In the first approach, the spatial dimension of the data is explored before the detection of temporal structures (SFTT). For the second approach (TFTS), the temporal dimension is first explored and then, the temporal features are used to detect spatial features. In the third approach (i.e. ParB), we use two parallel branches, one to extract spatial features and one to extract temporal features. The two sets of features are then combined before they are given to the classifier. In Fig. \ref{per_dataset}, we present the $F1score$, $AUCPR$, $AUROC$ and $PAI@5$, respectively for the DL approaches, compared to the 10 baseline approaches, per dataset for the ``All Crimes'' crime type in the highest resolution of $p=40$ i.e. $40\times40$ cells. The ROC-curves and PR-curves for the same experiment are shown in Fig. \ref{curves_per_crime}.
Similar behaviour appears for the rest of the crime types.

From the results we can see that all three DL approaches give consistently better performance than the baseline methods in the binary classification task of a cell being a hotspot or not, with SFTT being the winning approach. For each cell the SFTT approach gives the highest scores in all metrics for all datasets except the San Francisco one. The TFTS approach follows in the second place while the ParB approach does not perform well in this task.

\subsection{Evaluation of Cell Size}
\label{sec:res:eval_cellsize}

The second major parameter that we evaluated is the effect of spatial resolution on the approaches. The resolutions that were tested vary from the coarsest resolution of $p=16$ (i.e. $16\times16$ cells grid) to the finest of $p=40$ cells grid with a step of $p=8$ cells and the results for every metric are presented in Table \ref{tbl:percellresults}. While increasing the number of cells, the feature maps become sparser. 

\begin{table*}[!th]
\caption{F1score, Precision-Recall AUC, ROC AUC and PAI@5 for different $p$ (i.e. different resolutions) in the Philadelphia dataset for ``All Crimes'' crime type. The winning algorithm is in bold.}
\label{tbl:percellresults}
\resizebox{7.1in}{!}{%
\begin{tabular}{ l | c c c c | c c c c | c c c c | c c c c }
  & \multicolumn{4}{c|}{F1score} & \multicolumn{4}{c|}{ AUCPR} & \multicolumn{4}{c|}{AUROC} & \multicolumn{4}{c}{PAI@5} \\
Algorithm & 16 & 24 & 32 & 40 & 16 & 24 & 32 & 40 & 16 & 24 & 32 & 40 & 16 & 24 & 32 & 40 \\
\hline
CCRBoost & 0.93 & 0.92 & 0.88 & 0.88 & 0.99 & 0.99 & 0.98 & 0.97 & 0.92 & 0.92 & 0.90 & 0.91 & 1.87 & 1.94 & 1.86 & 2.00 \\
Decision Trees (C4.5) & 0.93 & 0.92 & 0.90 & 0.89 & 0.99 & 0.97 & 0.97 & 0.96 & 0.85 & 0.80 & 0.78 & 0.79 & 0.59 & 1.51 & 1.54 & 1.80 \\
Naive Bayes & 0.92 & 0.87 & 0.84 & 0.83 & 0.98 & 0.98 & 0.98 & 0.97 & 0.82 & 0.89 & 0.86 & 0.88 & 1.21 & 2.82 & 2.00 & 1.93 \\
Logit Boost & 0.94 & 0.92 & 0.91 & 0.89 & 1.00 & 0.99 & 0.99 & 0.99 & 0.99 & 0.96 & 0.94 & 0.94 & 0.81 & 1.73 & 2.52 & 1.91 \\
SVM & 0.94 & 0.93 & 0.91 & 0.90 & 0.99 & 0.99 & 0.98 & 0.98 & 0.91 & 0.93 & 0.90 & 0.90 & 0.06 & 0.12 & 0.19 & 0.25 \\
Random Forests & 0.94 & 0.92 & 0.90 & 0.89 & 1.00 & 0.99 & 0.98 & 0.98 & 0.97 & 0.94 & 0.91 & 0.92 & 1.04 & 2.44 & 1.93 & 1.98 \\
KNN & 0.94 & 0.92 & 0.91 & 0.89 & 0.99 & 0.98 & 0.97 & 0.97 & 0.86 & 0.87 & 0.84 & 0.88 & 0.93 & 1.66 & 1.89 & 2.29 \\
MLP (150) & 0.94 & 0.92 & 0.90 & 0.89 & 0.98 & 0.99 & 0.98 & 0.98 & 0.84 & 0.94 & 0.91 & 0.92 & 2.21 & 2.09 & 2.30 & 2.95 \\
MLP (150, 300, 150, 50) & 0.94 & 0.92 & 0.90 & 0.89 & 0.97 & 0.99 & 0.98 & 0.98 & 0.79 & 0.91 & 0.91 & 0.90 & 1.31 & 2.10 & 2.15 & 2.70 \\
ST-ResNet & 0.91 & 0.88 & 0.86 & 0.82 & 0.99 & 0.99 & 0.99 & 0.99 & 0.98 & 0.96 & 0.96 & 0.96 & 3.30 & 1.62 & 2.55 & 2.56 \\
SFTT & \textbf{0.99} & \textbf{0.97} & \textbf{0.96} & \textbf{0.94}  & \textbf{1.00} & \textbf{1.00} & \textbf{1.00} & \textbf{0.99} & \textbf{0.99} & \textbf{0.98} & \textbf{0.97} & \textbf{0.97} & 4.02 & 4.34 & \textbf{4.14} &\textbf{ 4.33} \\
TFTS & 0.99 & 0.97 & 0.95 & 0.94 & 1.00 & 0.98 & 0.99 & 0.99 & 0.95 & 0.81 & 0.93 & 0.96 & 4.30 & \textbf{4.40} & 4.10 & 4.12 \\
ParB & 0.99 & 0.96 & 0.94 & 0.92 & 0.99 & 0.99 & 0.99 & 0.99 & 0.88 & 0.91 & 0.94 & 0.94 & \textbf{4.32} & 3.41 & 3.29 & 3.31 \\
\hline
\end{tabular}
}
\end{table*}

It is evident that increasing the spatial resolution leads to a deterioration of $F1score$ for all models, including the baselines. 
On the other hand, the $PAI$ metric seems to benefit from the greater detail in some of the compared algorithms, which is expected, since the same proportion of crimes is correctly predicted but inside a smaller proportion of the study area.
In the AUC metrics, the SFTT approach constantly outperfoms all other approaches.

Having these results as guide to our further research, the SFTT approach was selected for the rest of our experiments as the winning configuration.

\subsection{Evaluation of Spatial Body}
\label{sec:res:eval_spatialBody}
\begin{figure}[!t]
\centering
\subfloat[]{
\includegraphics[width=3.3in, height=1.6in]{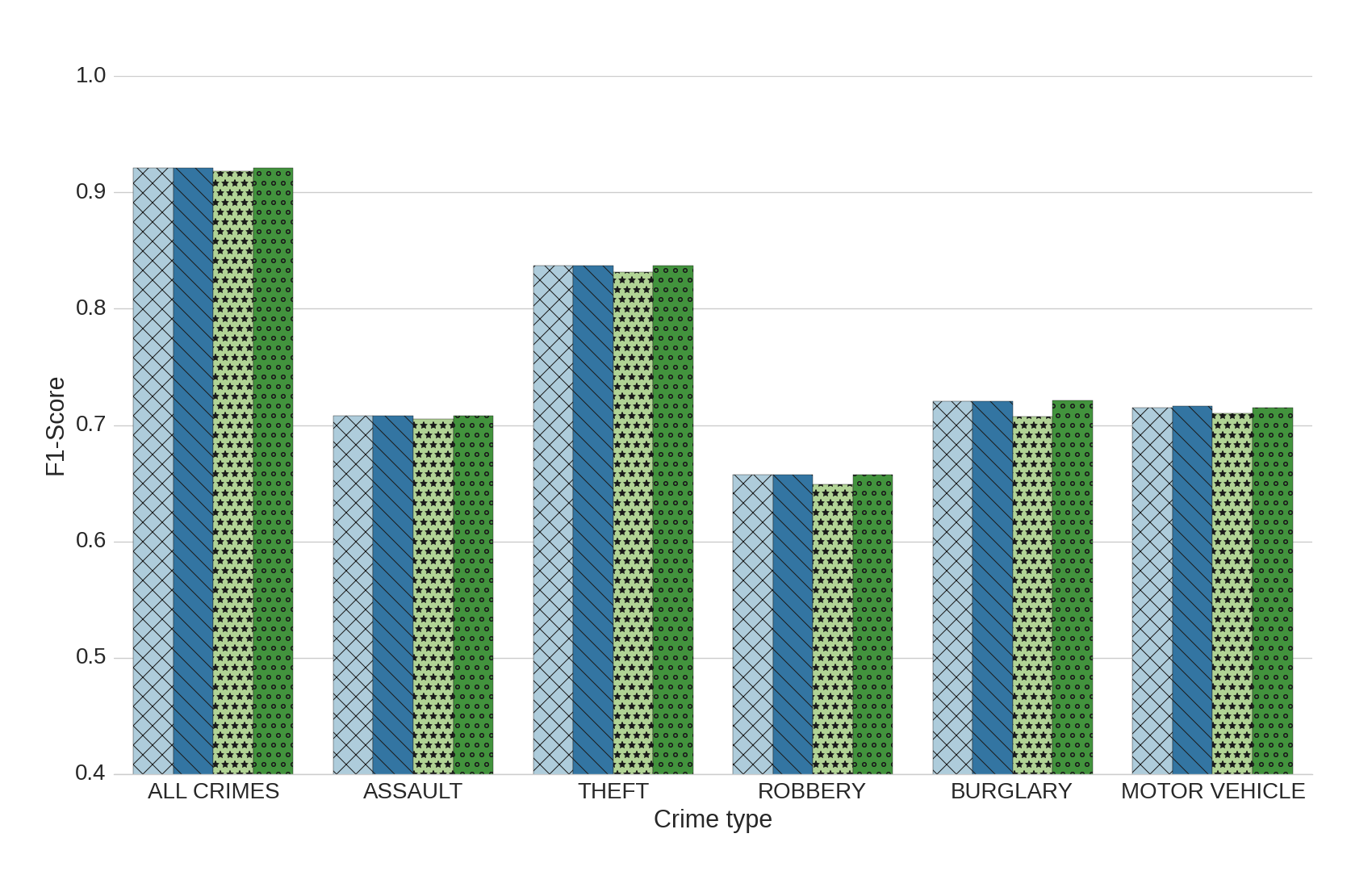}
\label{body_per_crime_f1score}
}\hfill
\subfloat[]{
\includegraphics[width=3.0in, height=1.6in]{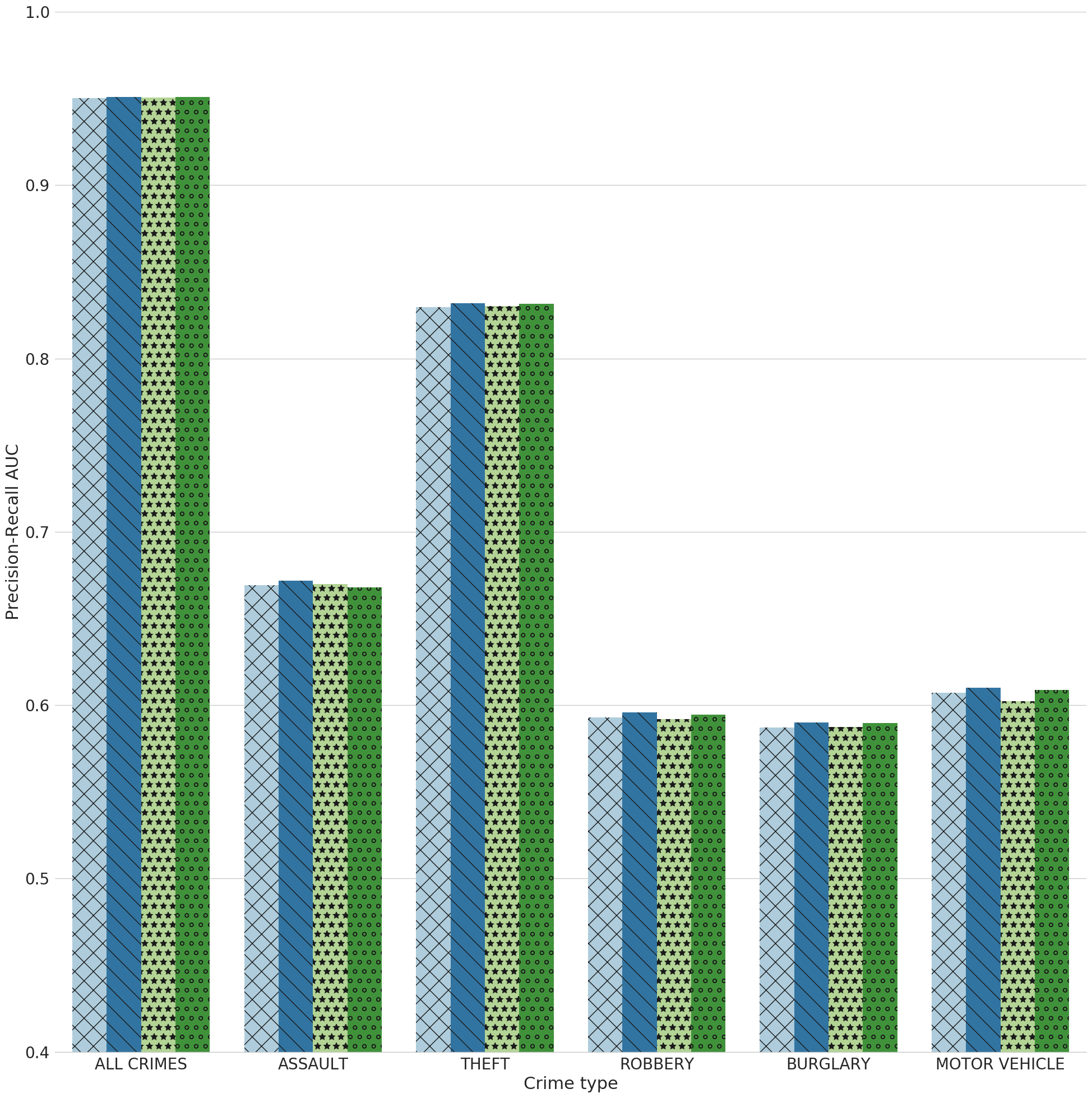}
\label{body_per_crime_pr_auc}
}\hfill
\subfloat[]{
\includegraphics[width=3.0in, height=1.6in]{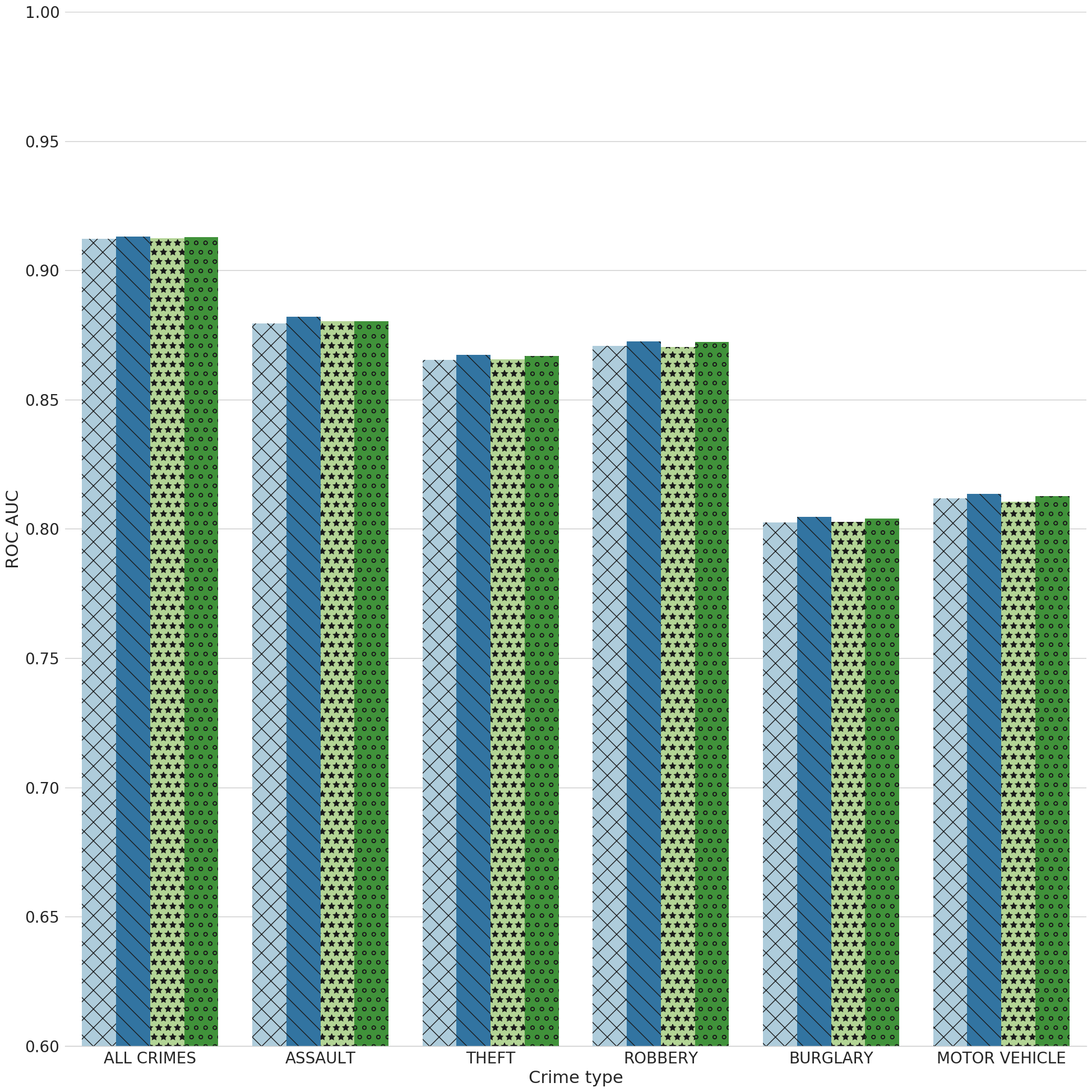}
\label{body_per_crime_roc_auc}
}\hfill
\subfloat[]{
\includegraphics[width=3.0in, height=1.6in]{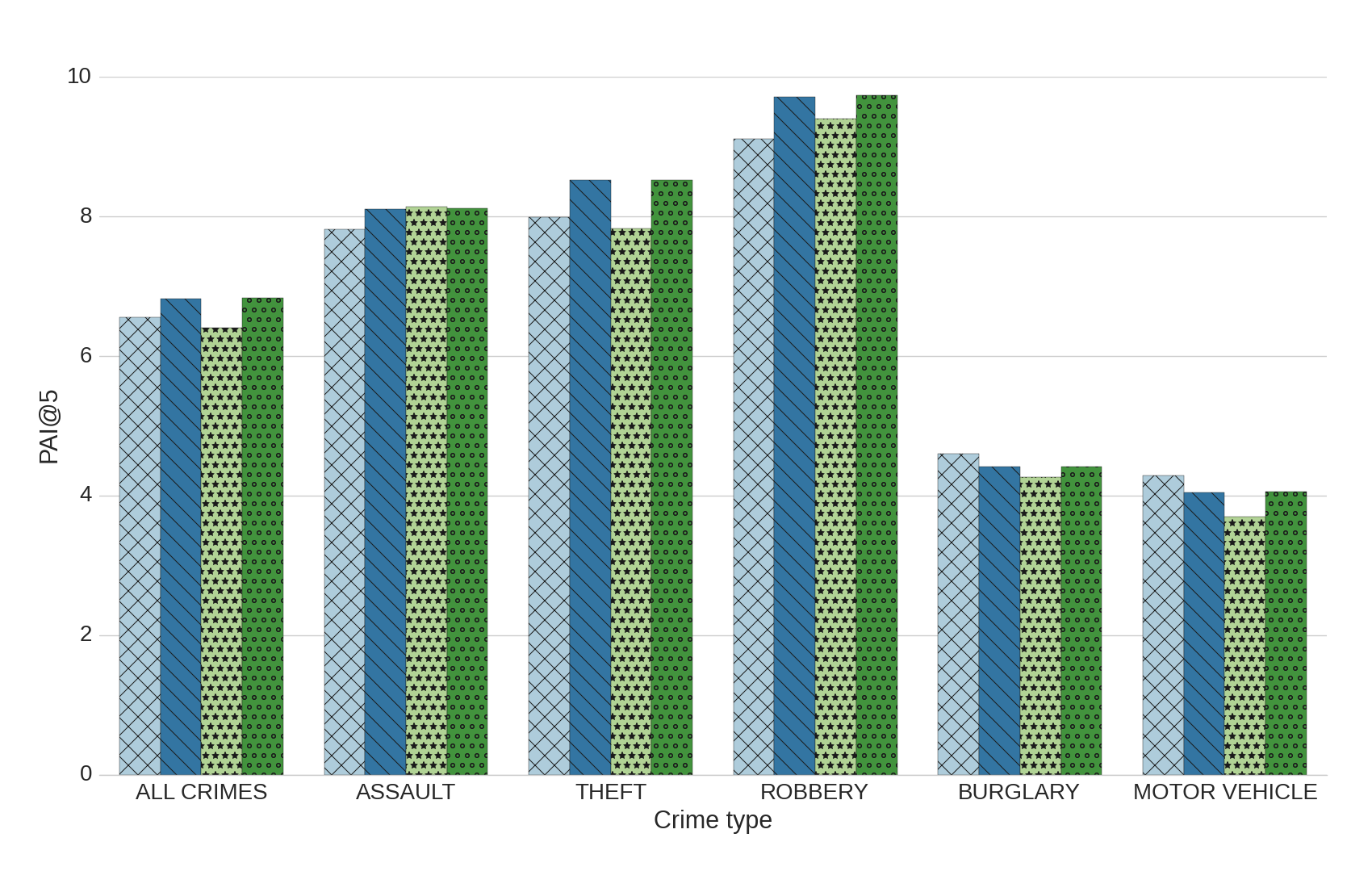}
\label{body_per_crime_pai5}
}\hfill
\subfloat{
\includegraphics[width=3.34in]{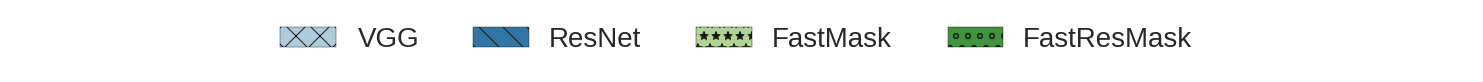}
\label{body_per_crime_legend}
}\hfill
\caption{(a) F1score, (b) PR AUC, (c) ROC AUC and (d) PAI@5 per crime type in the Philadelphia dataset, for 4 spatial clue extractor bodies. Best viewed in color.}
\label{body_per_crime}
\end{figure}

Having selected the SFTT approach, Fig. \ref{body_per_crime} presents a comparison of the 4 bodies' performance for various crime types. In all of the metrics the FastMask model has the worst performance compared to the other bodies. While the FastResMask body has similar performance with the ResNet body, the later has fewer parameters (therefore a smaller memory footprint) and is faster to train (Table \ref{tbl:complexity}). We can conclude that for this task, the extraction of features from different scales of the data is reduntant and ineffective.

The VGG body performs equivalently with the ResNet body in the F1score metric (Fig. \ref{body_per_crime_f1score}) which measures the correctly identified hotspots when considering all the cells inside the study area. In the AUC metrics (Fig. \ref{body_per_crime_pr_auc} and \ref{body_per_crime_roc_auc}), which measure the probability that a randomly selected hotspot is correctly identified as such, the VGG body performs better than the ResNet body.

On the other hand, in the 4 crime types with the more incidents, the performance of the VGG body is inferior to that of ResNet for the PAI metric (Fig. \ref{body_per_crime_pai5}) which measures if the more significant (in terms of more occuring crimes) cells are marked as more probable to be hotspots.

A new set of experiments was conducted for the selection and modeling of input data with only 1 crime category as input versus all 11 crime categories as input. Our experiments verified that the proposed setup of having all 11 crime categories as input channels contributes positively to the F1score performance of the model by a factor of $2.5\%$ in the Seattle dataset, up to $4.5\%$ in the Philadelphia dataset.

\subsection{Evaluation of Temporal Body}
\label{sec:res:eval_tempBody}
The quality of temporal clues extracted from the temporal part of our SFTT model can be affected by the depth and the width of the RNN. 
To evaluate these two parameters we performed experiments firstly by changing only the depth of LSTM layers in the network and then by changing only the width of the LSTM layers.  

The depth of the temporal part of the network is defined by the number of LSTM layers present in the network. For every consecutive LSTM layer a new level of non linear abstraction is introduced in the network. We evaluated the effect by using from one to nine LSTM layers to reveal that there was no significant gain by adding more layers. More specifically, the $F1score$ of our tested model in the Philadelphia ``All Crimes'', which is the class with most data, was $0.94289 \pm 0.00003$ in all examined depths. Following the same behaviour, the $F1score$ of the model on Philadelphia ``Burglary'' (sparse with few records) was at $ 0.76475 \pm 0.00017$ for all numbers of LSTM layers.

The width of the network is defined by the number of hidden states that exist in each LSTM layer and defines the amount of temporal information that can be modelled. To examine the effect that LSTM layers' width has on the performance of the model, we varied the width up to 500 hidden neurons in each LSTM layer. We observed in the results that almost all the available temporal information can be captured with LSTM layers of only 50 hidden neurons.

\subsection{Evaluation of Batch Normalization and Dropout}
\label{sec:res:eval_batchNormDropout}
The amount of data that is available for the training process is fairly small, compared to image classification datasets with millions of samples. This lack of data can lead to networks that overfit by learning the training data but not being able to generalise properly. 

In DL literature the two most common methods to overcome overfitting problems is the use of Batch Normalization and Dropout. We used Batch Normalization after every pooling layer and tested several variations of Dropout layers. When we removed Batch Normalization the performance dropped significantly.

A Dropout layer, randomly zeroes out a percentage of the input data values, during the training phase, thus creating variations of the training data in each epoch. These random variations effectively multiply the number of training samples and allow the networks to generalize better. By increasing the percentage of dropout, more virtual samples are created but the data become sparser. 
In order to find the best compromise, we investigated dropout percentages from 10\% up to 90\%. Adding more dropout helps the predictor to generalize a little bit better for the more populated ``All Crimes'' crime type in predicting the most probable hotspots. For all other crime types, varying the percentage of dropout has no measurable effect.

\subsection{Multi-label hotspot classification}
\label{sec:res:eval_multi}

\begin{figure}[!t]
\centering
\subfloat{
	\includegraphics[width=2.8in, height=1.5in]{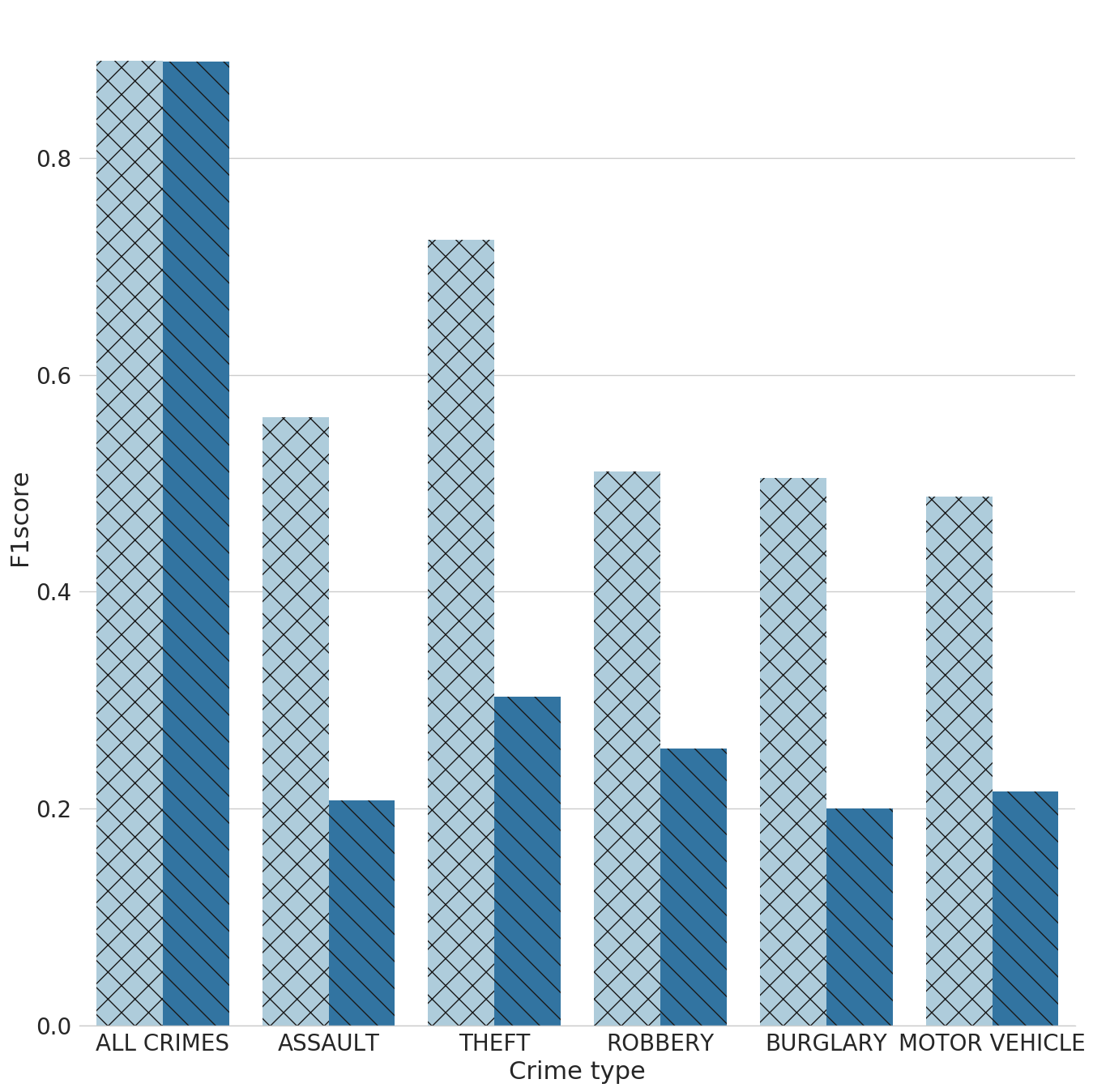}
	\label{multiclass_fmeasure}
}\hfill
\subfloat{
	\includegraphics[width=3.3in]{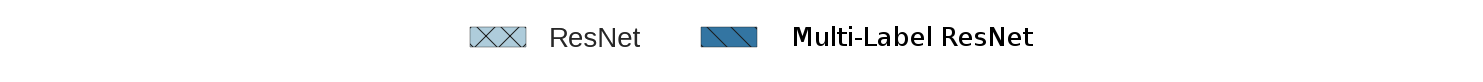}
}
\hfill
\caption{F1score per crime type in the Philadelphia dataset, for binary and multi-label hotspot prediction. Best viewed in color.}
\label{multiclass}
\end{figure}

%
%
As a final experiment on the configuration of the proposed approach, a multi-label classification was tested. The aim was to evaluate how the proposed model performs when simultaneously predicting the probability for each cell to be a hotspot for all possible crime types at the same time. To do so, the last layers' dimension was changed from 1 (binary classification) to 11, which is the number of crime types in the dataset. The loss function was also modified from binary cross entropy to multi-class cross entropy. In order for the predictions of this setup to remain comparable with our previous results, we kept the last layer activation function to sigmoid, which is the typical for the configuration of multi-label models.
For this experiment, the SFTT model with ResNet spatial body was used and the results are presented in Fig. \ref{multiclass}. As it is expected, the binary classification model outperforms the multi-label model in each crime type except the ``All Crimes'' type were the same performance is observed. 
This is observed due to class imbalance as well as difficulty of the model to predict accurately all the labels (i.e. classes), due to the sparsity of the data.

\section{Conclusions and Future Work}
\label{conclusions}

In this paper we investigated the capability of DL methods to forecast hotspot areas in an urban environment, where crimes of certain types are more likely to occur in a defined future window. To achieve this goal we fed the DL methods with the minimum amount of data containing only spatial, temporal and crime type information. In order for the models to better predict the order of ``hotness'' we used a dual output setting where the second output is the number of crimes that occurred in the same future window.
Moreover, we selected our SFTT model configuration as the winning one, compared with 10 different algorithms in 5 crime incidence datasets and further analysed the selected parameters for robust results.  

In the future, we will investigate if incorporating additional information to our system, like temporal semantics, demographics, weather, street maps 
and points of interest in the area can help our model to learn better features. Temporal semantics can help the prediction of fluctuations in crime rates that depend on seasonal events, like holidays, and time of day events, like shift change in stores.

By replacing dense layers with convolutional in both the temporal clues extraction and the classification parts of the model, in a full convolutional network fashion, we can increase the spatial resolution. Another road that we did not explore is to pre-process and augment the available data. Preprocessing steps could include normalization of the data and blind source separation, while data augmentation can come from flipping and rotating the data on their spatial dimensions. Additional data could be created by changing the shift of the temporal sliding window from daily to hourly rate.



\printbibliography

\end{document}